\documentclass{article}
\pdfoutput=1 

\PassOptionsToPackage{numbers, compress}{natbib}

\usepackage[final]{neurips_2022}

\usepackage[utf8]{inputenc} %
\usepackage[T1]{fontenc}    %
\usepackage{hyperref}       %
\usepackage{url}            %
\usepackage{booktabs}       %
\usepackage{amsfonts}       %
\usepackage{nicefrac}       %
\usepackage{microtype}      %
\usepackage{xcolor}         %
\usepackage{enumitem} 

\usepackage{graphicx}
\usepackage{multirow}
\usepackage{subcaption}
\usepackage{algorithm}
\usepackage{algorithmic}

\newcommand{\TBD}[1]{{\bf \color{blue} To be continued...}}

\usepackage{amsmath}
\DeclareMathOperator*{\argmax}{\bf arg\,max}
\DeclareMathOperator*{\argmin}{\bf arg\,min}
\newcommand{\vw}{\ensuremath{\mathbf w}}
\newcommand{\vv}{\ensuremath{\mathbf v}}

\usepackage[labelfont=bf]{caption} 
\title{Blackbox Attacks via Surrogate Ensemble Search}

\author{%
Zikui Cai,~~ Chengyu Song,~~ Srikanth Krishnamurthy, \\ {\bf Amit Roy-Chowdhury,~~ M. Salman Asif\thanks{Corresponding authors: Zikui Cai (\texttt{zcai032@ucr.edu}) and M. Salman Asif (\texttt{sasif@ucr.edu})}} \\
University of California Riverside\\
}

\begin{document}

\maketitle

\begin{abstract}
  Blackbox adversarial attacks can be categorized into  transfer- and query-based attacks. Transfer methods do not require any feedback from the victim model, but provide lower success rates compared to query-based methods. Query attacks often require a large number of queries for success. To achieve the best of both approaches, recent efforts have  tried to combine them, but still require hundreds of queries to achieve high success rates (especially for targeted attacks). In this paper, we propose a novel method for Blackbox Attacks via Surrogate Ensemble Search (BASES) that can generate highly successful blackbox attacks using an extremely small number of queries. We first define a perturbation machine that generates a perturbed image by minimizing a weighted loss function over a fixed set of surrogate models. To generate an attack for a given victim model, we search over the weights in the loss function using queries generated by the perturbation machine. Since the dimension of the search space is small (same as the number of surrogate models), the search requires a small number of queries. We demonstrate that our proposed method achieves better success rate with at least $30\times$ fewer queries compared to state-of-the-art methods on different image classifiers trained with  ImageNet (including VGG-19, DenseNet-121, and ResNext-50). In particular, our method requires as few as 3 queries per image (on average) to achieve more than a $90\%$ success rate for targeted attacks and 1--2 queries per image for over a $99\%$ success rate for untargeted attacks. Our method is also effective on Google Cloud Vision API and achieved a $91\%$ untargeted attack success rate with 2.9 queries per image. We also show that the perturbations generated by our proposed method are highly transferable and can be adopted for hard-label blackbox attacks. Furthermore, we argue that BASES can be used to create attacks for a variety of tasks and show its effectiveness for attacks on object detection models. Our code is available at  {\color{magenta} \url{https://github.com/CSIPlab/BASES}}. 
\end{abstract}

\section{Introduction}
\begin{figure}[t]
\centering
    \centering
    \includegraphics[width=0.95\textwidth]{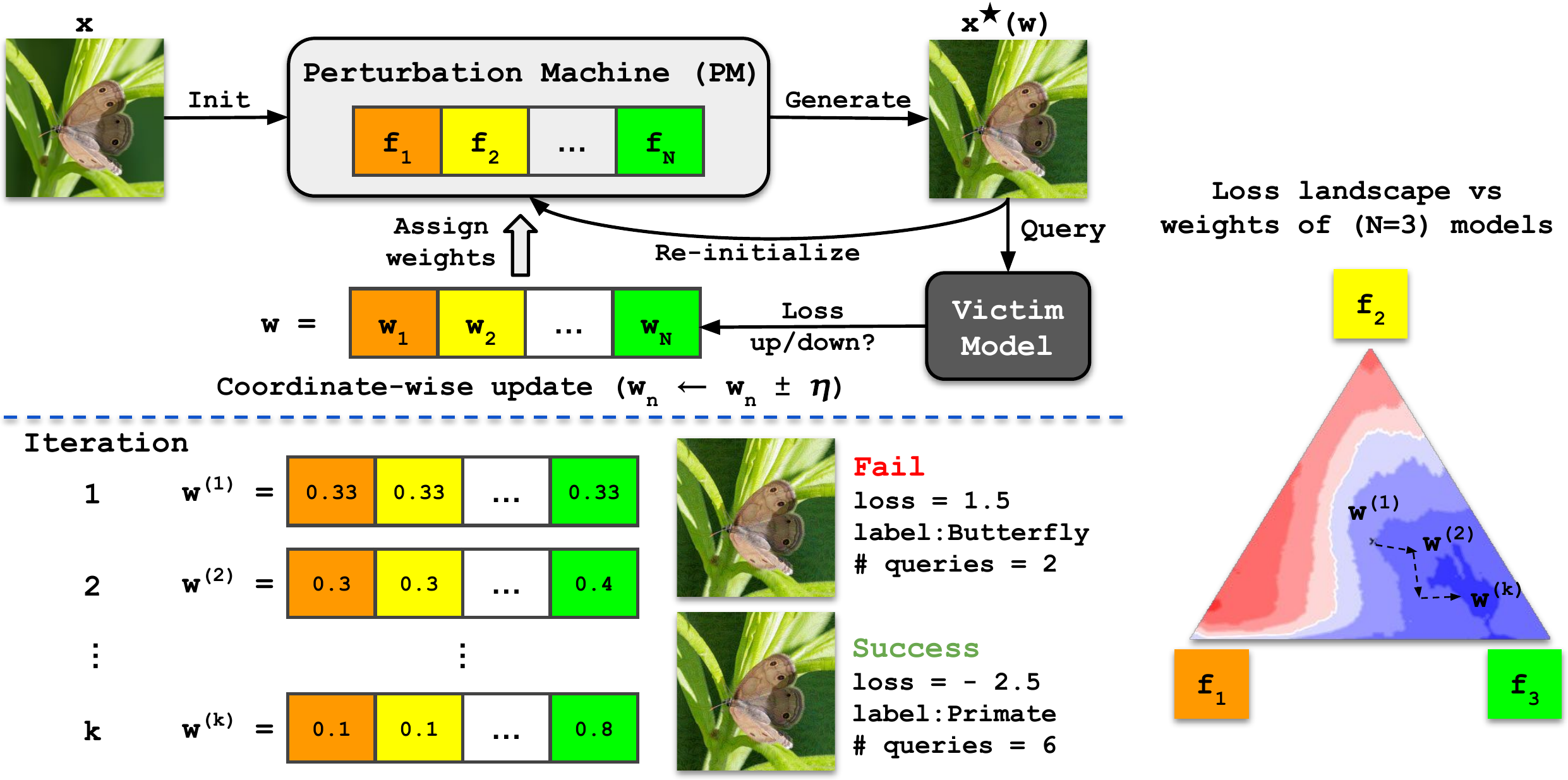}
\caption{BASES for score-based attack. (\textbf{Top-left}) We define a perturbation machine (PM) using a fixed set of $N$ surrogate models, each of which is assigned a weight value as $\mathbf{w} = [w_1,\ldots,w_N]$. The PM generates a perturbed image $x^\star(\vw)$ for a given input image $x$ by minimizing the perturbation loss that is defined as a function of $\mathbf{w}$. To fool a victim model, we update one coordinate in $\mathbf{w}$ at a time while querying the victim model using $x^\star(\vw)$ generated by the PM. We can view this approach as a bi-level optimization or search procedure; the PM generates a perturbed image with the given weights $x^\star(\vw)$ in the inner level, while we update $\mathbf{w}$ in the outer level. (\textbf{Bottom-left}) We visualize weights and perturbed images for a few iterations. We stop as soon as the attack is successful (e.g. original label - `Butterfly' is changed to target label - `Primate' for targeted attack). (\textbf{Right}) Victim loss values for different weights along the Barycentric coordinates on the triangle. We start with equal weights (at the centroid) and traverse the space of $\mathbf{w}$ to reduce loss (concentrate on model {$f_3$}). Red color indicates large loss values (unsuccessful attack), and blue indicates low loss (successful attack).} \label{fig:intro}
\end{figure}
Deep neural network (DNN) models are known to be vulnerable to adversarial attacks \cite{szegedy2013intriguing,goodfellow2014explaining,papernot2016transferability,papernot2017practical}. Many methods have been proposed in recent years to generate adversarial attacks  \cite{goodfellow2014explaining,kurakin2016adversarial,madry2017towards,dong2018boosting,xie2019improving,lin2019nesterov,huang2019black,lord2022attacking} (or to defend against such attacks  \cite{madry2017towards,tramer2017ensemble,xu2017feature,guo2017countering,meng2017magnet,liu2018towards,samangouei2018defense,xie2019feature,ren2020adversarial,bai2021recent}). 
Attack methods for blackbox models 
can be divided into two broad categories: transfer- and query-based methods. Transfer-based methods generate attacks for some (whitebox) surrogate models via backpropagation and test if they fool the victim models \cite{papernot2016transferability,papernot2017practical}. 
They are usually agnostic to victim models as they do not require or readily use any feedback; and they often provide lower success rates compared to query-based methods. %
On the other hand, query-based attacks achieve high success rate but at the expense of querying the victim model several times to find perturbation directions that reduce the victim model loss \cite{chen2017zoo,tu2019autozoom,guo2019simple,li2020qeba,ilyas2018black}.
One possible way to achieve a high success rate while keeping the number of queries small, is to combine the transfer and query attacks. While there has been impressive recent work along this direction \cite{cheng2019improving,huang2019black,tashiro2020diversity,lord2022attacking}, the state-of-the-art methods \cite{tashiro2020diversity,lord2022attacking} still require hundreds of or more queries to be successful at targeted attacks. Such attacks are infeasible for limited-access settings where a user cannot query a model that many times\cite{goodfellow2019research}.

Given this premise, we design a new method for blackbox attacks via surrogate ensemble search (BASES), combining transfer and query ideas, to fool a given victim model with higher success rates and fewer queries compared to state-of-the-art methods. 
For example, our evaluation shows that BASES (on average) only requires 3 queries per image to achieve over 90\% success rate for targeted attacks, which is  at least $30\times$ fewer queries compared to state-of-the-art methods \cite{huang2019black,lord2022attacking}.
BASES consists of two key steps that can be viewed as bilevel optimization steps. 1) A perturbation machine generates a query for the victim model based on weights assigned to the surrogate models. 2) The victim model's feedback is used to change weights of the perturbation machine to refine the query. Figure~\ref{fig:intro} depicts these steps.

We first define a perturbation machine (PM) that generates a single perturbation to fool all the (whitebox) models in the surrogate ensemble. We use a surrogate ensemble for two reasons: 
1) It is known to provide better transfer attacks \cite{liu2017delving,dong2018boosting}. The assumption is that if an adversarial image can fool multiple surrogate models, then it is very likely to fool a victim model as well. For the same reason, an ensemble with different and diverse surrogate models provides better attack transfer. 
2) Our main interest is in searching for perturbations that can fool the given victim model. A single surrogate model provides a fixed perturbation; hence, it does not offer flexibility to search over perturbations. To facilitate search over perturbations, we define the adversarial loss for the PM as a function of weights assigned to each model in the ensemble. By changing the weights of the loss function, we can generate different perturbations and steer in a direction that fools the victim model. 
It is worth noting that perturbations generated by a surrogate ensemble with an arbitrary set of weights often fools all the surrogate models, but they do not guarantee success on unseen victim models; therefore, searching over the weights space for surrogate models is necessary. 

Since the number of models in the surrogate ensemble is small, the search space is low dimensional and requires extremely small number of queries compared to other query-based approaches.
In our method, we further simplify the search process by updating one weight element at a time, which is equivalent to coordinate descent, which has been shown to be effective in query-based attacks \cite{chen2017zoo,guo2019simple}. 
Since we search along one coordinate at a time instead of estimating the full gradients, the method is extremely efficient in terms of query count. %
In particular, our method requires two queries per coordinate update but offers success rates as good as that given by performing a full gradient update step (as shown in Section~\ref{sec:experiments}). 
Reducing the dimension of the search space while maintaining high success rate for query-based attacks is an active area of research \cite{guo2019simple,huang2019black,tashiro2020diversity,lord2022attacking}, and our proposed method pushes the boundary in this area.

We perform extensive experiments for (score-based) blackbox attacks using a variety of surrogate and blackbox victim models for both targeted and untargeted attacks. We select PyTorch Torchvision \cite{paszke2017automatic} as our model zoo, which contains 56 image classification models trained on ImageNet \cite{deng2009imagenet} that span a wide range of architectures. We demonstrate superior performance by a large margin over state-of-the-art approaches, especially for targeted attacks. Furthermore, we tested the perturbations generated by our method for attacks on hard-label classifiers. Our results show that the perturbations generated by our method are highly transferable. We present also present experiments for attacks on object detectors in the supplementary material, which demonstrate the effectiveness of our attack method for tasks beyond image classification.

The main contributions of this paper are as follows.
\begin{itemize}[topsep=0pt, leftmargin=*, itemsep=1pt]
    \item We propose a novel, yet simple method, BASES, for  effective and query-efficient blackbox attacks. %
    The method adjusts weights of the surrogate ensemble by querying the victim model and achieves high fooling rate targeted attack with a very small number of queries.
    \item We perform extensive experiments to demonstrate that BASES outperforms state-of-the-art methods \cite{cheng2019improving,huang2019black,tashiro2020diversity,lord2022attacking} by a large margin; over 90\% targeted success rate with less than 3 queries, which is at least $30\times$ fewer queries than other methods.
    \item We also demonstrate the effectiveness under a real-world blackbox setting by attacking Google Cloud Vision API and achieve 91\% untargeted fooling rate with 2.9 queries ($3\times$ less than \cite{huang2019black}).
    \item The perturbations from BASES are highly transferable and can also be used for hard-label attacks. In this challenging setting, we can achieve over 90\% fooling rate for targeted and almost perfect fooling rate for untargeted attacks on a variety of models using less than 3 and 2 queries, respectively.
    \item We show that BASES can be used for different tasks by creating attacks for object detectors that significantly improve the fooling rate over transfer attacks.
\end{itemize}

\section{Related work}

\noindent\textbf{Ensemble-based transfer attacks.} 
Transferable adversarial examples that can fool one model can also fool a different model \cite{papernot2016transferability,papernot2017practical,liu2017delving,li2020towards}. Transfer-based untargeted attacks are considered `easy' since the adversarial examples can disrupt feature extractors into unrelated directions (e.g., in MIM \cite{dong2018boosting}, the fooling rate for some models can be as high as $87.9\%$). In contrast, transfer-based targeted attacks often suffer from low fooling rates (e.g., MIM shows a transfer rate of about $20\%$ at best). To improve the transfer rate, several methods use ensemble based approach. To combine the information from different surrogate models;  \cite{liu2017delving} fuses probability scores and \cite{dong2018boosting} proposes combining logits. While these methods have been effective, the most natural and generic approach is to combine losses, which can be used for tasks beyond classification \cite{che2020new,wu2020making,cai2022context}. MGAA \cite{yuan2021meta} iteratively selects a set of surrogate models from an ensemble and performs meta train and meta test steps to reduce the gap between whitebox and blackbox gradient directions. Simulator-Attack method \cite{ma2021simulating} uses several surrogate models to train a generalized substitute model as a "Simulator" for the victim model; however, training such simulators is computationally expensive and  difficult to scale to large datasets.
Previous ensemble approaches typically assign equal weights for each surrogate model. In contrast, we update weights for different surrogate models based on the victim model feedback.

\noindent\textbf{Query-based attacks.}
Unlike transfer-based attacks, query-based attacks do not make assumptions that surrogate models share similarity with the victim model. They can usually achieve high fooling rates even for targeted attacks (but at the expense of queries) \cite{chen2017zoo,ilyas2018black,tu2019autozoom}. The query complexity is proportional to the dimension of the search space. Queries over the entire image space can be extremely expensive \cite{chen2017zoo}, requiring millions of queries for targeted attack \cite{tu2019autozoom}. 
To reduce the query complexity, a number of approaches have attempted to reduce the search space dimension or leverage transferable priors or surrogate models to generate queries \cite{guo2019simple,huang2019black,tashiro2020diversity,suya2019hybrid, li2021adversarial}. SimBA-DCT \cite{guo2019simple} searches over the low DCT frequencies. P-RGF \cite{cheng2019improving} utilizes surrogate gradients as a transfer-based prior, and draws random vectors from a low-dimensional subspace for gradient estimation. TREMBA \cite{huang2019black} trains a perturbation generator and traverses over the low-dimensional latent space. ODS \cite{tashiro2020diversity} optimizes in the logit space to diversify perturbations for the output space. GFCS \cite{lord2022attacking} searches along the direction of surrogate gradients, and falls back to ODS if surrogate gradients fail. Some other methods \cite{suya2019hybrid,yang2020learning,ma2021simulating} also reuse the query feedback to update surrogate models or `blackbox simulator', but such a fine-tuning process provides very slight improvements. We summarize the typical search space and average number of queries for some state-of-the-art methods in Table~\ref{tab:pros-cons}.
In our approach, we further shrink the search dimension to as low as the number of models in the ensemble. Since our search space is dense with adversarial perturbations, we show that a moderate-size ensemble with $20$ models can generate successful targeted attacks for a variety of victim models while requiring only 3 queries (on average), which is at least $30$ time fewer than that of existing methods.

\section{Method}
\label{sec:method}
\subsection{Preliminaries}

We use additive perturbation \cite{szegedy2013intriguing,goodfellow2014explaining,madry2017towards} to generate a perturbed image as $x^\star = x+ \delta$, where $\delta$ denotes the perturbation vector of same size as input image $x$. To ensure that the perturbation is imperceptible to humans, we usually constrain its $\ell_p$ norm to be less than a threshold, i.e., $\|\delta\|_p \leq \varepsilon$, where $p$ is usually chosen from $\{2, \infty\}$.
Such adversarial attacks for a victim model $f$ can be generated by minimizing the so-called adversarial loss function $\mathcal{L}$ over $\delta$ such that the output $f(x+\delta)$ is as close to the desired  (adversarial) output as possible. Specifically, the attack generator function maps the input image $x$ to an adversarial image $x^\star$ such that the output $f(x^\star)$ is either far/different from the original output $y$ for untargeted attacks, or close/identical to the desired output $y^\star$ for targeted attacks.

Let us consider a multi-class classifier $f(x): x \mapsto z$, where $z = [z_1, \ldots, z_C]$ represents a logit vector at the last layer. The logit  vector can be converted to a probability vector $p = \text{softmax}(z)$. We refer to such a classifier as a ``score-based'' or ``soft-label'' classifier. 
In contrast, a ``hard-label'' classifier provides a single label index out of a total of $C$ classes. 
We can derive the hard label from the soft labels as $y = \argmax_{c} f(x)_c$. 
For untargeted attacks, the objective is to find $x^\star$ such that $\argmax_{c}f(x^\star)_c \neq y$. For targeted attacks, the objective is to find $x^\star$ such that $\argmax_{c}f(x^\star)_c = y^\star$, where $y^\star$ is the target label.

Many efforts on  adversarial attacks use iterative variants of the fast signed gradient method (FGSM)  \cite{goodfellow2014explaining} because of their simplicity and effectiveness. Notable examples include I-FGSM \cite{kurakin2016adversarial}, PGD \cite{madry2017towards}, and MIM \cite{dong2018boosting}. We use PGD attack in our PM, which iteratively optimizes perturbations as
\begin{equation}\label{eq:PGD}
    \delta^{t+1} = \Pi_{\varepsilon} \left( \delta^{t} - \lambda~ \mathbf{sign} (\nabla_{\delta}\mathcal{L}(x+\delta^{t}, y^\star)) \right),
\end{equation}
where $\mathcal{L}$ is the loss function and $\Pi_{\varepsilon}$ denotes a projection operator. There are many loss functions suitable for crafting adversarial attacks. We mainly employ the following margin loss, which has been shown to be effective in C\&W attacks \cite{carlini2017towards}:
\begin{equation}\label{eq:loss-cw}
    \mathcal{L}(f(x), y^\star) = \max \left(\max _{j \neq {y^\star}}f(x)_j - f(x)_{y^\star},-\kappa\right),
\end{equation}
where $\kappa$ is the margin parameter that adjusts the extent to which the example is `adversarial.' A larger $\kappa$ corresponds to a lower optimization loss. One advantage of C\&W loss function is that its sign directly indicates whether the attack is successful or not ($+ve$ value indicates failure, $-ve$ value indicates success). Cross-entropy loss is also a popular loss function to consider, which has similar performance as margin loss (comparison results are provided in the supplementary material).

\subsection{Perturbation machine with surrogate ensemble} 

\noindent\textbf{Controlled query generation with PM.} We define a perturbation machine (PM) to generate queries for the victim model as shown in Figure~\ref{fig:intro}. The PM accepts an image and generates a perturbation to fool all the surrogate models. Furthermore, we seek some control over the perturbations generated by the PM to steer them in a direction that fools the victim model. To achieve these goals, we construct the PM such that it minimizes a weighted adversarial loss function over the surrogate ensemble. 

\noindent\textbf{Adversarial loss functions for ensembles.} Suppose our PM consists of $N$ surrogate models given as $\mathcal{F} = \{f_1,\ldots, f_N\}$, each of which is assigned a weight in $\vw = [w_1, \ldots, w_N]$ such that $\sum_{i=1}^N w_i = 1$. For any given image $x$ and $\vw$, we seek to find a perturbed image $x^\star(\vw)$ that fools the surrogate ensemble. Below we discuss three possible weighted ensemble loss functions-based optimization problems for targeted attack. Loss functions for untargeted attack can be derived similarly.
\begin{align}
    \textbf{weighted probabilities} \qquad x^\star(\vw) &= \argmin_x -\log{(\mathbf{1}_{y^\star} \cdot \sum_{i=1}^N w_i\, \text{softmax}(f_i(x)))}, \label{eq:fuse-prob} \\
    \textbf{weighted logits} \qquad x^\star(\vw) &=\argmin_{x} ~ \mathcal{L}(\sum_{i=1}^N w_if_i(x),y^\star), \label{eq:fuse-logit}\\
    \textbf{weighted loss} \qquad x^\star(\vw) &= \argmin_{x} ~ \sum_{i=1}^N w_i\mathcal{L}(f_i(x),y^\star).  \label{eq:fuse-loss} 
\end{align}
${y^\star}$ denotes the target label and $\mathbf{1}_{y^\star}$ denotes its one-hot encoding. $\mathcal{L}$ represents some adversarial loss function (e.g., C\&W loss).
The first problem in \eqref{eq:fuse-prob} is the minimization of the softmax cross-entropy loss defined on the weighted combination of probability vectors from all the models in the ensemble \cite{liu2017delving}.
The second problem in \eqref{eq:fuse-logit} optimizes adversarial loss over a weighted combination of logits from the models \cite{dong2018boosting}.
The third problem in \eqref{eq:fuse-loss} optimizes a weighted combination of adversarial losses over all models. The weighted loss formulation is the simplest and most generic ensemble approach that works not only for the classification task with logit or probability vectors, but also other tasks (e.g., object detection, segmentation) as long as the model losses can be aggregated \cite{wu2020making}. Here, we  focus on the weighted loss formulation, since it shows superior performance compared to weighted probabilities and logits formulations in our experiments (additional experiments are presented in the supplementary material).

Algorithm~\ref{alg:perturbation-machine} presents a pseudocode for the PM module for a fixed set of weights. The PM accepts an image $x$ and weights $\vw$ along with the surrogate ensemble and returns the perturbed image $x^\star = x+\delta$ after a fixed number of signed gradient descent steps (denoted as $T$) for the ensemble loss.

\begin{algorithm}[h] 
\caption{Perturbation Machine: $\delta,x^\star(\vw) = \textbf{PM}(x,\vw,\delta_\text{init})$}
\label{alg:perturbation-machine} 
\begin{algorithmic}[1]
\REQUIRE ~~\\
 Input $x$ and the target class $y^\star$ (for untargeted attack $y^\star \ne y$ ); \\
 Surrogate ensemble $\mathcal{F} = \{f_1, f_2, ..., f_N\}$; \\
 Ensemble weights $\vw=\{w_1, w_2, ..., w_N\}$; \\ 
 Initial perturbation $\delta_\text{init}$; Step size $\lambda$; Perturbation norm $(\ell_2/\ell_\infty)$ and bound $\varepsilon$; \\
 Number of signed gradient steps $T$ 
\ENSURE Adversarial perturbation $\delta, x^\star(\vw)$
\STATE $\delta = \delta_\text{init}$
\FOR{$ t = 1$ to $T$}
    \STATE Calculate $\mathcal{L}_\mathbf{ens} = \sum_{i=1}^N w_i\mathcal{L}_i(x+\delta, y^\star)$ \hfill {\color{magenta} \it $\triangleright$ Ensemble loss}
    \STATE Update $\delta \gets \delta - \lambda \cdot \mathbf{sign} (\nabla_{\delta}\mathcal{L}_\mathbf{ens})$
    \hfill {\color{magenta} \it  $\triangleright$ Gradient of ensemble via backpropagation}
    \STATE Project $\delta \gets \Pi_{\varepsilon}(\delta)$ \hfill {\color{magenta} \it $\triangleright$ Project to the feasible set of $\ell_\infty$ or $\ell_2$ ball}
\ENDFOR
\STATE $x^\star(\vw) \gets x+\delta$
\RETURN $\delta,x^\star(\vw)$
\end{algorithmic}
\end{algorithm}

\subsection{Surrogate ensemble search as bilevel optimization} 

Let us assume that we are given a blackbox victim model, $f_\vv$, that we seek to fool using a perturbed image generated by the PM (as illustrated in Figure~\ref{fig:intro}). 
Suppose the adversarial loss for the victim model is defined as $\mathcal{L}_\vv$.
To generate a perturbed image that fools the victim model, we want to solve the following optimization problem: 
\begin{equation}\label{eq:bilevel}
    \vw = \argmin_\vw \; \mathcal{L}_\vv(f_\vv(x^\star(\vw)),y^\star). 
\end{equation}
The problem in \eqref{eq:bilevel} is bilevel optimization that seeks to update the weight vector $\vw$ for the PM so that the generated $x^\star(\vw)$ fools the victim model. 
The PM in Algorithm~\ref{alg:perturbation-machine} can be viewed as a function that solves the inner optimization problem in our bilevel optimization. The outer optimization problem searches over $\vw$ to steer the PM towards a perturbation that fools the victim model.

\noindent\textbf{BASES: Blackbox Attacks via Surrogate Ensemble Search.}
Our objective is to maximize the attack success rate and minimize the number of queries on the victim model; hence, we adopt a simple yet effective iterative procedure to update the weights $\vw$ and generate a sequence of queries. Pseudocode for our  approach is shown in Algorithm~\ref{alg:bi-level}. 
We initialize all entries in $\vw$ to $1/N$ and generate the initial perturbed image $x^\star(\vw)$ for input $x$. We stop if the attack succeeds for the victim model; otherwise, we update $\vw$ and generate a new set of perturbed images. 
We follow \cite{chen2017zoo} and update $\vw$ in a coordinate-wise manner, where at every outer iteration, we select $n$th index and generate two instances of $\vw$ as $\vw^+,\vw^-$ by updating $w_n$ as $w_n+\eta,w_n-\eta$, where $\eta$ is a step size.
We normalize the weight vectors so that the entries are non-negative and add up to 1. We generate perturbations $x^\star(\vw^+),x^\star(\vw^-)$ using the PM and query the victim model. We compute the victim loss (or score) for $\{\vw^+,\vw^-\}$ and select the weights, the perturbation vector, and the perturbed images corresponding to the smallest victim loss. We stop if the attack is successful with any query. 

\begin{algorithm}[h] 
\caption{\textbf{BASES}: Blackbox Attack via Surrogate Ensemble Search}
\label{alg:bi-level}
\begin{algorithmic}[1]
\REQUIRE ~~\\
 Input $x$ and the target class $y^\star$ (for untargeted attack $y^\star \ne y$ ); \\
 Victim model $f_\vv$; Maximum number of queries $Q$; Learning rate $\eta$; \\
 Perturbation machine (PM) with surrogate ensemble
\ENSURE Adversarial perturbation $\delta, x^\star$
\\
\STATE Initialize $\delta = 0$; $q = 0$; $\vw = \{\nicefrac{1}{N}, \nicefrac{1}{N}, ..., \nicefrac{1}{N}\}$
\STATE Generate perturbation via PM: $\delta, x^\star(\vw) = \mathbf{PM}(x, \vw, \delta)$ \hfill  {\color{magenta} \it $\triangleright$ first query with equal weights}
\STATE Query victim model: $z = f_\vv(x+\delta)$
\STATE Update query count: $q\gets q+1$
\IF {$\argmax_{c}z_c = y^\star$}
    \STATE \textbf{break} \hfill {\color{magenta} \it $\triangleright$ stop if attack is successful}
\ENDIF
\WHILE{$ q < Q$} %
        \STATE Update surrogate ensemble weights as follows. \hfill  {\color{magenta} \it $\triangleright$ outer level updates weights}
        \STATE Pick a surrogate index $n$ \hfill  {\color{magenta} \it $\triangleright$ cyclic or random order}
        \STATE Compute $\vw^+, \vw^-$ by updating $w_n$ as $w_n + \eta, w_n-\eta$, respectively 
        \STATE Generate perturbation $x^\star(\vw^+), x^\star(\vw^-)$ via PM \hfill {\color{magenta} \it $\triangleright$ inner level generates query}
        \STATE Query victim model: $f_\vv(x^\star(\vw^+)), f_\vv(x^\star(\vw^-))$ \hfill {\color{magenta} \it $\triangleright$  2 queries per coordinate} 
        \STATE Calculate victim model loss for $\{\vw^+,\vw^-\}$ as $\mathcal{L}_\vv(\vw^+), \mathcal{L}_\vv(\vw^-)$ 
        \STATE Select $\vw, \delta, x^\star(\vw)$ for the weight vector with the smallest loss 
        \STATE Increment $q$ after every query, and stop if the attack is successful for any query
\ENDWHILE
\RETURN $\delta$
\end{algorithmic}
\end{algorithm}

\section{Experiments}\label{sec:experiments} 

\subsection{Experiment setup}
In this section, we present experiments on attacking the image classification task. Additional experiments on attacking object detection task can be found in the supplementary material.

\textbf{Surrogate and victim models.} We present experiments for blackbox attacks mainly using pretrained image classification models from Pytorch Torchvision \cite{paszke2017automatic}, which is a comprehensive and actively updated package for computer vision tasks. 
At the time of writing this paper, Torchvision offers $56$ classification models trained on ImageNet dataset \cite{deng2009imagenet}. These models have different architectures and include the family of VGG~\cite{simonyan2014very}, ResNet~\cite{he2016deep}, SqueezeNet~\cite{iandola2016squeezenet}, DenseNet\cite{huang2017densely}, ResNeXt~\cite{xie2017aggregated}, MobileNet~\cite{sandler2018mobilenetv2,howard2019searching},  EfficientNet~\cite{tan2019efficientnet}, RegNet~\cite{radosavovic2020designing}, Vision Transformer~\cite{dosovitskiy2020image}, and ConvNeXt~\cite{liu2022convnet}.
We choose different models as the victim blackbox models for our experiments, as shown in Figures \ref{fig:compare-sota-linf-targeted}, \ref{fig:compare-wb-bb}, and \ref{fig:hard-label}.
To construct an effective surrogate ensemble for the PM, we sample 20 models from different families: 
\texttt{\{VGG-16-BN, ResNet-18, SqueezeNet-1.1, GoogleNet, MNASNet-1.0, DenseNet-161, EfficientNet-B0, RegNet-y-400, ResNeXt-101, Convnext-Small, ResNet-50, VGG-13, DenseNet-201, Inception-v3, ShuffleNet-1.0, MobileNet-v3-Small, Wide-ResNet-50, EfficientNet-B4, RegNet-x-400, VIT-B-16\}}.
We vary our ensemble size $N \in \{4,10,20\}$ by picking the first $N$ model from the set. In most of the experiments, our method uses $N=20$ models in the \textbf{PM}, unless otherwise specified.
We also tested a different set of models pretrained on TinyImageNet dataset, the details of which are included in the supplementary material.
To validate the effectiveness of our methods in a practical blackbox setting, we also tested Google Cloud Vision API.

\textbf{Comparison with other methods.} We compare our method  with some of the state-of-the-art methods for score-based blackbox attacks. TREMBA \cite{huang2019black} is a powerful attack method that searches for perturbations by changing the latent code of a generator trained using a set of surrogate models. 
GFCS \cite{lord2022attacking} is a recently proposed surrogate-based attack method that probes the victim model using the surrogate gradient directions. We use their original code repositories \cite{code-gfcs,code-tremba}. For completeness, we also compare with two earlier methods, ODS \cite{tashiro2020diversity} and P-RGF \cite{cheng2019improving}, that leverage transferable priors, even though they have been shown to be less effective than GFCS and TREMBA.
Additional details about comparison with TREMBA and GFCS are provided in the supplementary material. 

\textbf{Dataset.} We mainly use 1000 ImageNet-like images from the NeurIPS-17 challenge \cite{neurips17challenge,kurakin2018adversarial}, which provides the ground truth label and a target label for each image. We also provide evaluation results on TinyImageNet in the supplementary material.

\textbf{Query budget.} In this paper, we move towards a limited-access setting, since for many real-life applications, legitimate users will not be able to run many queries \cite{goodfellow2019research}. In contrast with TREMBA and GFCS, which set the maximum query count to $10,000$ and $50,000$, respectively, we set the maximum count to be $500$ and only run our method for $50$ queries in the worst case. (TREMBA also uses only 500 queries for Google Cloud Vision API to cut down the cost.) 

\textbf{Perturbation budget.} We evaluated our method under both $\ell_\infty$ and $\ell_2$ norm bound, with commonly used perturbation budgets of $\ell_\infty \leq 16$ and $\ell_2 \leq 255\sqrt{0.001D} = 3128$ on a 0--255 pixel intensity scale, where $D$ denotes the number of pixels in the image. For attacking Google Cloud Vision API, we reduce the norm bound to $\ell_\infty \leq 12$ to align with the setting in TREMBA. Results for $\ell_2$ norm bound are provided in the supplementary material. 

\textbf{Targeted vs untargeted attacks.} All the methods achieve near perfect fooling rates for untargeted attacks in our experiments. This is because untargeted attack on image classifiers is not challenging \cite{lord2022attacking}, especially when the number of classes is large. Thus, we primarily report experimental results on targeted attacks in the main text and report results for untargeted attacks in the supplementary material. We use the target labels provided in the dataset \cite{neurips17challenge} in the experiments discussed in the main text. We provide additional analysis on using different target label selection methods such as `easiest' and `hardest' according to the confidence scores in the supplementary material.

\subsection{Score-based attacks}

\begin{figure}[t]
\centering
\begin{subfigure}[c]{0.3\linewidth}
    \centering
    \includegraphics[width=0.95\textwidth]{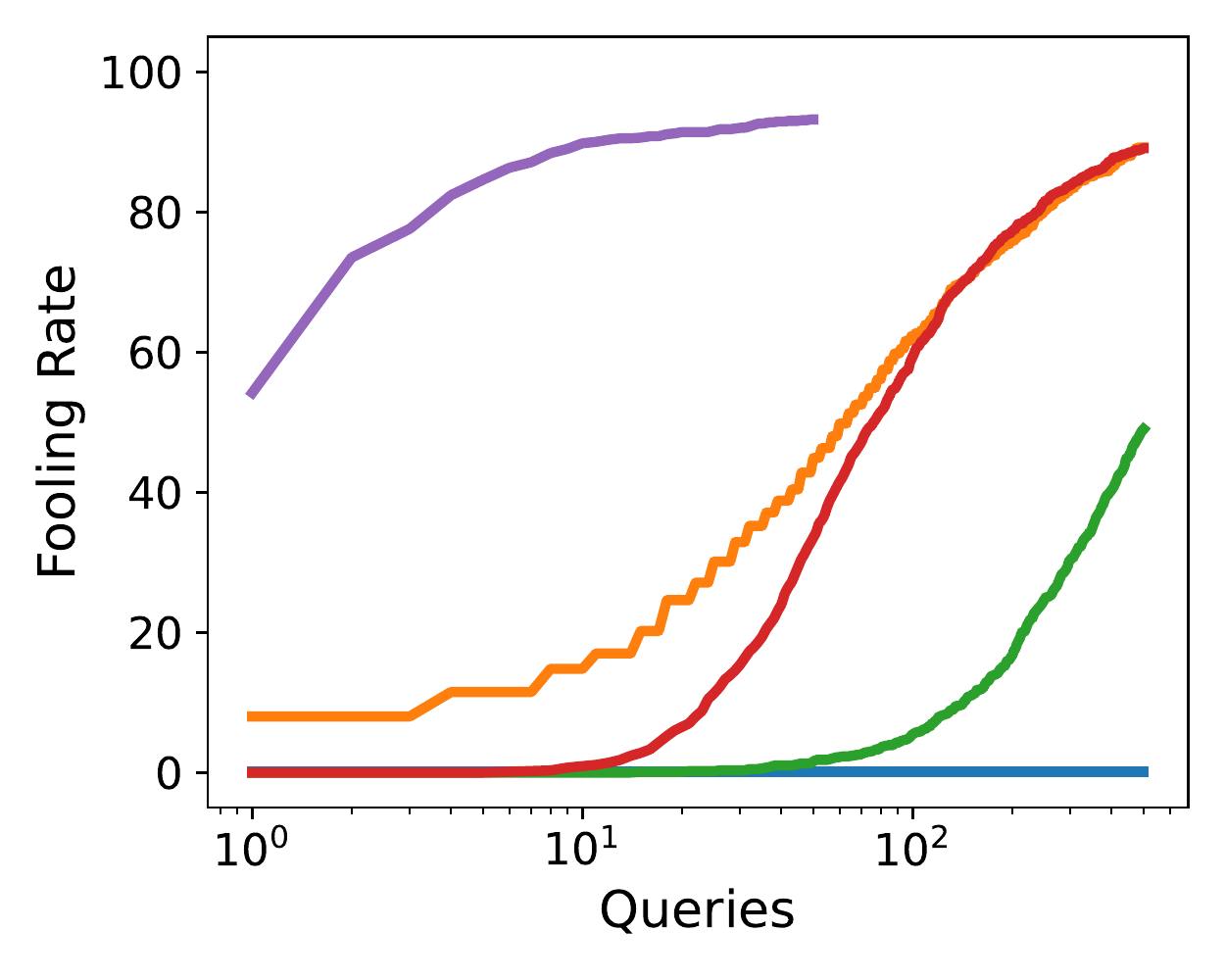}
    \caption{VGG-19}
    \label{fig:compare-sota-linf-targeted-vgg19}
\end{subfigure}
\begin{subfigure}[c]{0.3\linewidth}
    \centering
    \includegraphics[width=0.95\textwidth]{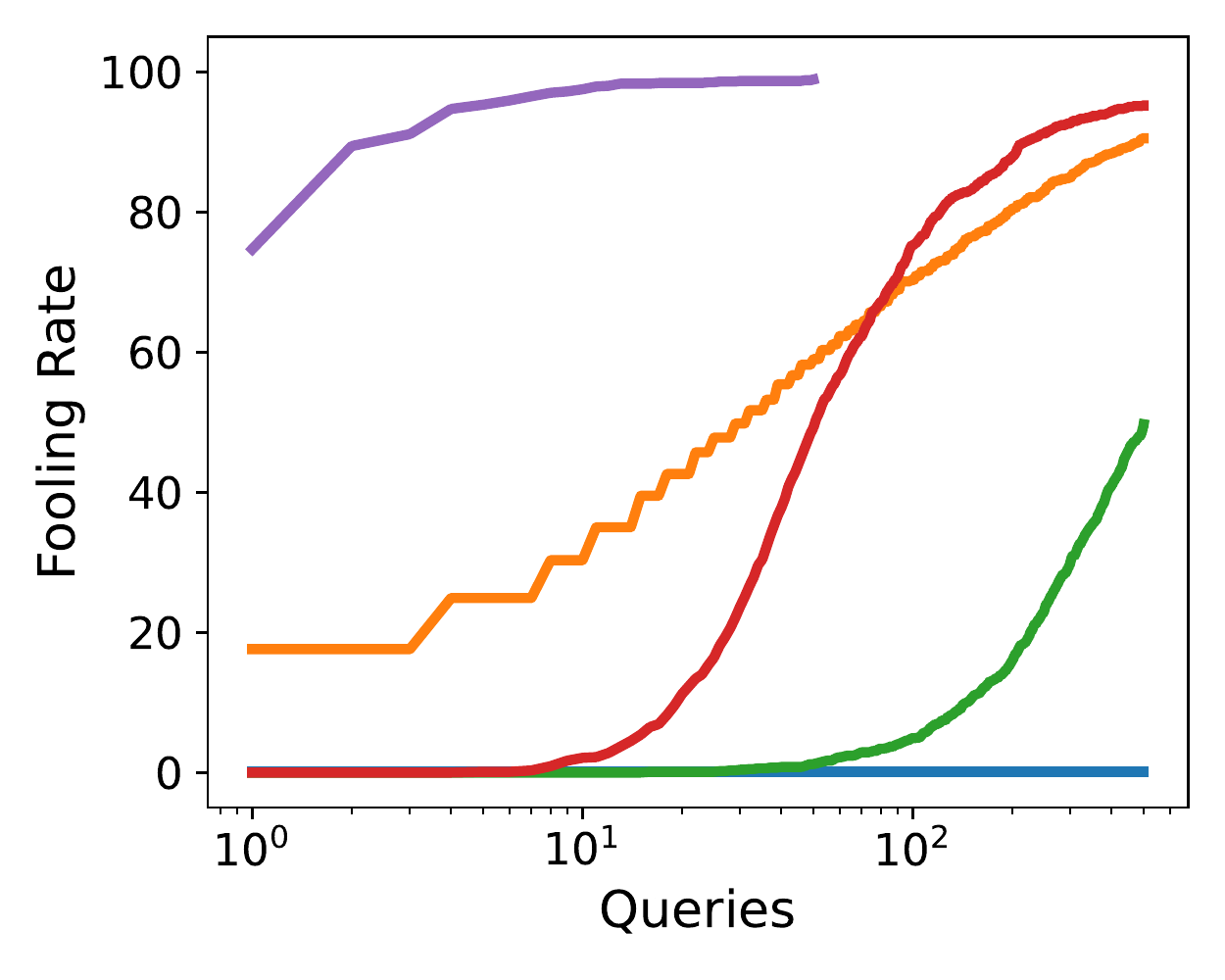}
    \caption{DenseNet-121}
\end{subfigure}
\begin{subfigure}[c]{0.38\linewidth}
    \centering
    \includegraphics[width=0.99\textwidth]{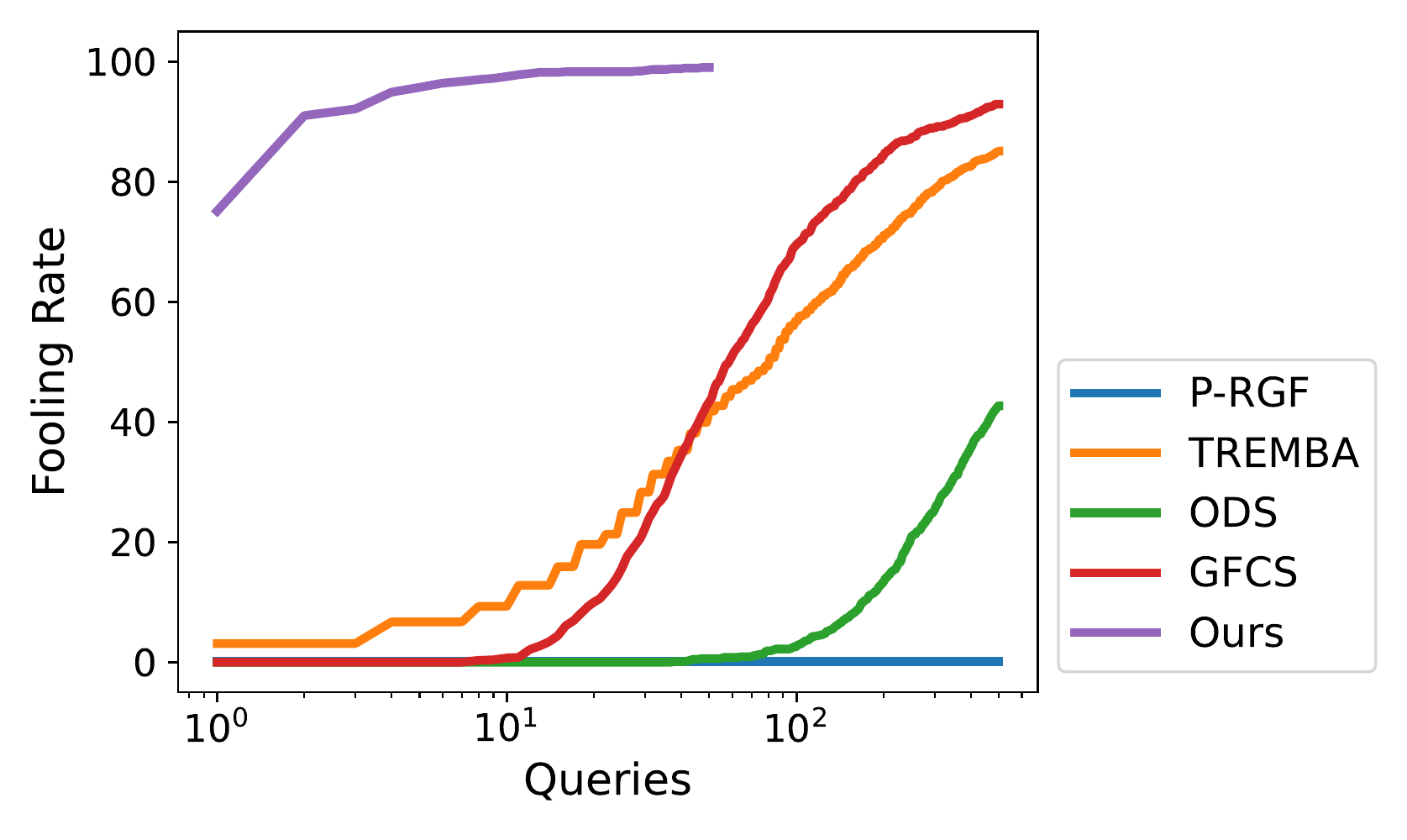}
    \caption{ResNext-50}
\end{subfigure}
\caption{Comparison of 5 attack methods on three victim models under perturbation budget $l_\infty \leq 16$ for targeted attack. Our method achieves high success rate (over 90\%) with few queries (average of 3  per image).}
\label{fig:compare-sota-linf-targeted}
\end{figure}

\noindent\textbf{Targeted attacks.} 
Figure~\ref{fig:compare-sota-linf-targeted} presents a performance comparison of five methods for targeted attacks on three blackbox victim models. Our proposed method provides the highest fooling rate with the least number of queries. P-RGF is found to be ineffective (almost $0\%$ success) for targeted attacks under low query budgets. TREMBA and GFCS are similar in performance; TREMBA shows better performance when query count is small, but GFCS matches TREMBA after nearly $100$ queries. Nevertheless, our method clearly outperforms these two powerful methods by a large margin at any level of query count. 
We summarize the search space dimension $\mathcal{D}$ and query counts vs fooling rate of different methods under a limited (and realistic) query budget for both the targeted and untargeted attacks in Table \ref{tab:pros-cons}. Our method is the most effective in terms of fooling rate vs number of queries (and has the smallest search dimension).  \textit{Additional results and details about fair comparison and fine tuning of TREMBA and GFCS are provided in the supplementary material. }

\begin{table}[t]
\centering
\small
\caption{Number of queries vs fooling rate of different methods and the search space dimension $\mathcal{D}$.}
\label{tab:pros-cons}
\begin{tabular}{lccccccc}
    \hline
    \multirow{3}{*}{Method} & \multirow{3}{*}{$\mathcal{D}$} & \multicolumn{6}{c}{Number of queries ($\mathbf{mean}\pm\mathbf{std}$) per image and fooling rate} \\
     &  & \multicolumn{2}{c}{VGG-19} & \multicolumn{2}{c}{DenseNet-121} & \multicolumn{2}{c}{ResNext-50} \\ \cline{3-8} 
     &  & Targeted & Untargeted & Targeted & Untargeted & Targeted & Untargeted \\ \hline
    \vspace{0.1cm}
    P-RGF \cite{cheng2019improving} & 7,500 & - & \begin{tabular}[c]{@{}c@{}}156 $\pm$ 113\\  93.5\%\end{tabular} & - & \begin{tabular}[c]{@{}c@{}}164 $\pm$ 112\\ 92.9\%\end{tabular} & - & \begin{tabular}[c]{@{}c@{}}166 $\pm$ 116\\ 92.5\%\end{tabular} \\ 
    \vspace{0.1cm}
    TREMBA \cite{huang2019black} & 1,568 & \begin{tabular}[c]{@{}c@{}}92 $\pm$ 107\\ 89.2\%\end{tabular} & \begin{tabular}[c]{@{}c@{}}2.4 $\pm$ 14\\ 99.7\%\end{tabular} & \begin{tabular}[c]{@{}c@{}}70 $\pm$ 104\\ 90.5\%\end{tabular} & \begin{tabular}[c]{@{}c@{}}5.9 $\pm$ 28\\ 99.5\%\end{tabular} & \begin{tabular}[c]{@{}c@{}}100 $\pm$ 109\\ 85.1\%\end{tabular} & \begin{tabular}[c]{@{}c@{}}7.5 $\pm$ 38\\ 98.9\%\end{tabular} \\ 
    \vspace{0.1cm}
    ODS \cite{tashiro2020diversity} & 1,000 & \begin{tabular}[c]{@{}c@{}}261 $\pm$ 125\\ 49.0\%\end{tabular} & \begin{tabular}[c]{@{}c@{}}38 $\pm$ 48\\ 99.9\%\end{tabular} & \begin{tabular}[c]{@{}c@{}}266 $\pm$ 123\\ 49.7\%\end{tabular} & \begin{tabular}[c]{@{}c@{}}52 $\pm$ 64\\ 99.0\%\end{tabular} & \begin{tabular}[c]{@{}c@{}}270 $\pm$ 116\\ 42.7\%\end{tabular} & \begin{tabular}[c]{@{}c@{}}54 $\pm$ 65\\ 98.4\%\end{tabular} \\ 
    \vspace{0.1cm}
    GFCS \cite{lord2022attacking} & 1,000 & \begin{tabular}[c]{@{}c@{}}101 $\pm$ 95\\ 89.1\%\end{tabular} & \begin{tabular}[c]{@{}c@{}}14 $\pm$ 21\\ 100.0\%\end{tabular} & \begin{tabular}[c]{@{}c@{}}76 $\pm$ 75\\ 95.2\%\end{tabular} & \begin{tabular}[c]{@{}c@{}}16 $\pm$ 36\\ 99.9\%\end{tabular} & \begin{tabular}[c]{@{}c@{}}86 $\pm$ 87\\ 92.9\%\end{tabular} & \begin{tabular}[c]{@{}c@{}}15 $\pm$ 18\\ 99.7\%\end{tabular} \\
    \textbf{Ours} & \textbf{20} & \textbf{\begin{tabular}[c]{@{}c@{}}3.0 $\pm$ 5.4\\ 95.9\%\end{tabular}} & \textbf{\begin{tabular}[c]{@{}c@{}}1.2 $\pm$ 2.4\\ 99.8\%\end{tabular}} & \textbf{\begin{tabular}[c]{@{}c@{}}1.8 $\pm$ 2.7\\ 99.4\%\end{tabular}} & \textbf{\begin{tabular}[c]{@{}c@{}}1.2 $\pm$ 1.8\\ 99.9\%\end{tabular}} & \textbf{\begin{tabular}[c]{@{}c@{}}1.8 $\pm$ 2.6\\ 99.7\%\end{tabular}} & \textbf{\begin{tabular}[c]{@{}c@{}}1.2 $\pm$ 0.9\\ 100.0\%\end{tabular}} \\ \hline
\end{tabular}
\end{table}

\begin{figure}[h]
\centering
\begin{subfigure}[c]{0.3\linewidth}
    \centering
    \includegraphics[width=0.95\textwidth]{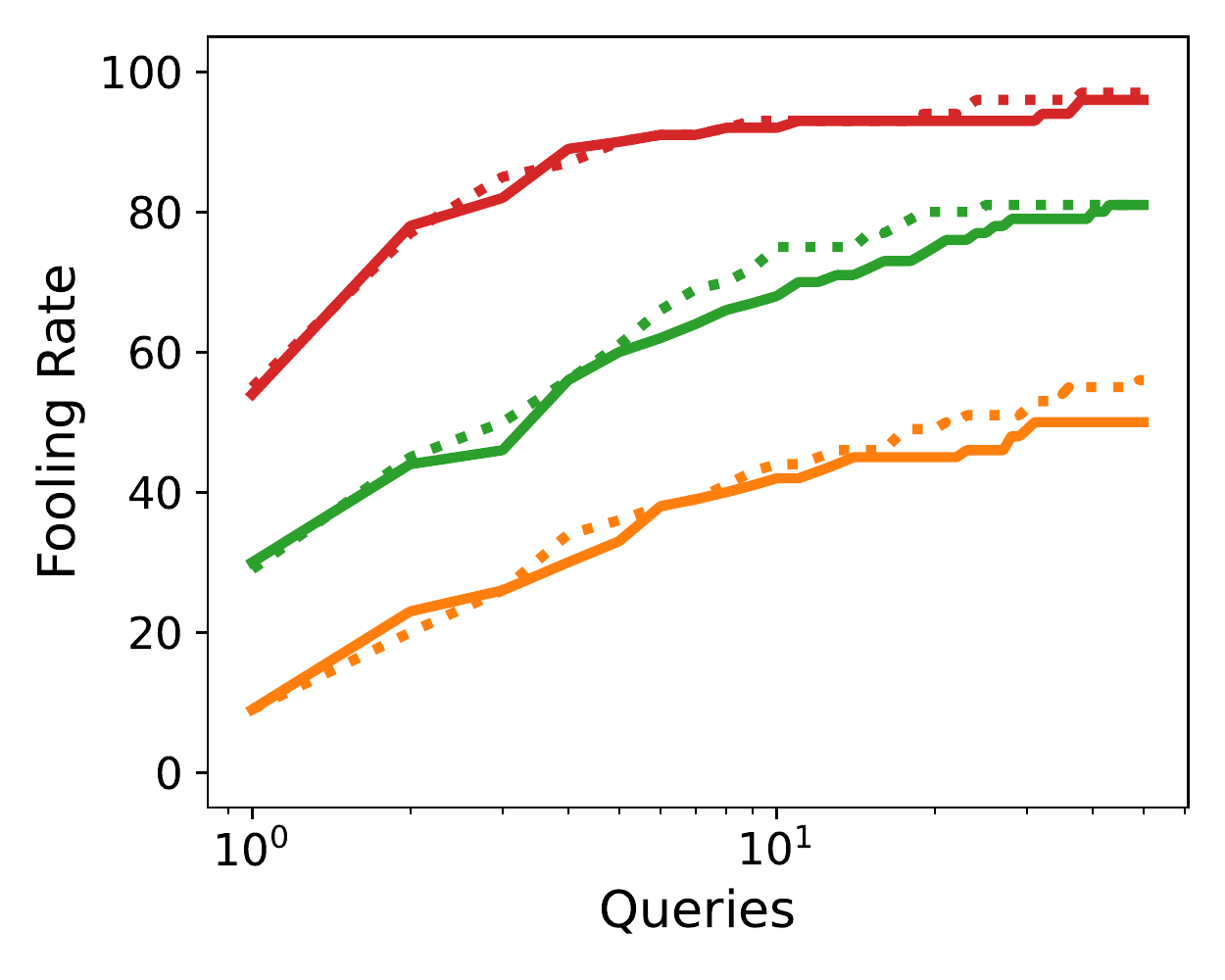}
    \caption{VGG-19}
\end{subfigure}
\begin{subfigure}[c]{0.3\linewidth}
    \centering
    \includegraphics[width=0.95\textwidth]{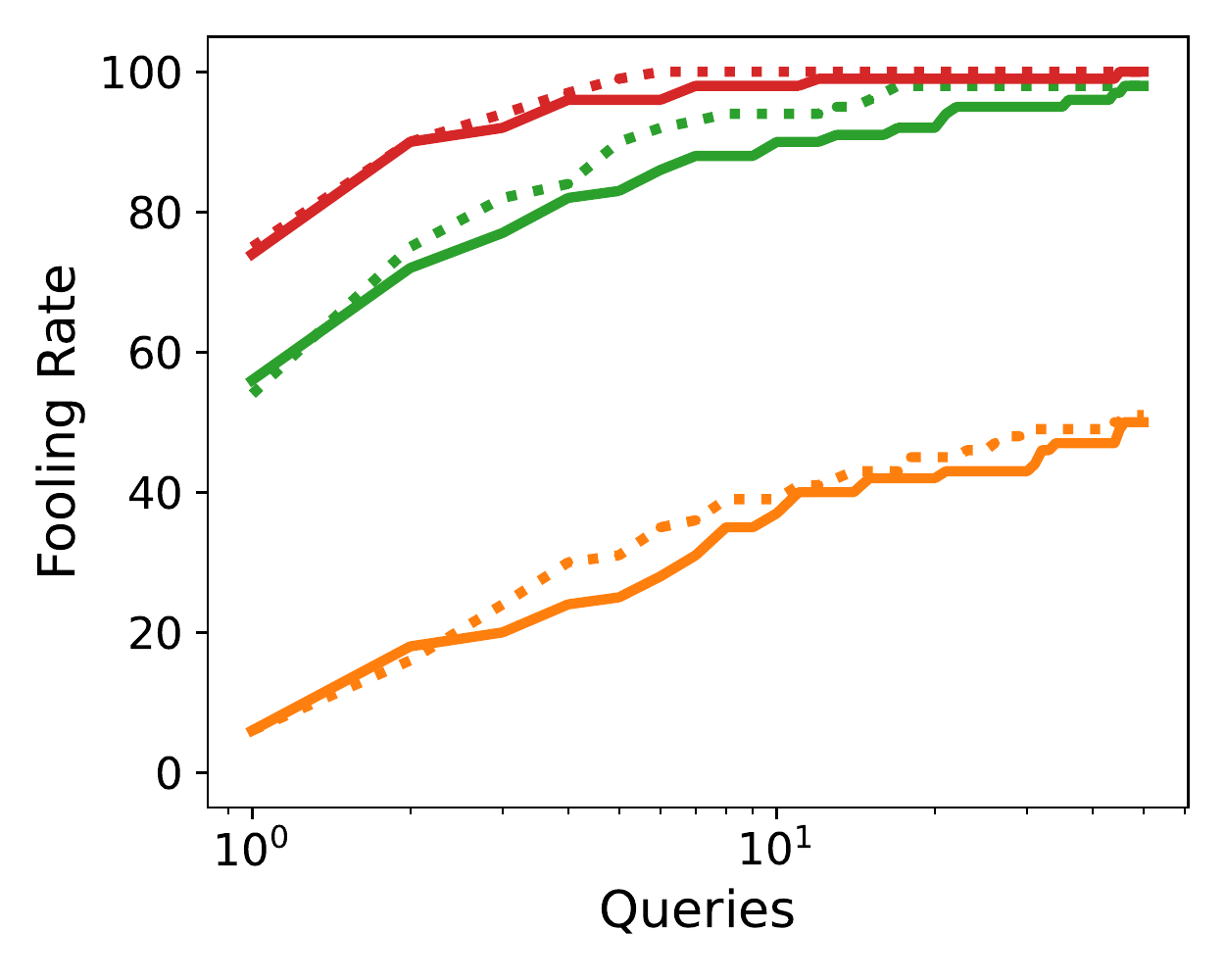}
    \caption{DenseNet-121}
\end{subfigure}
\begin{subfigure}[c]{0.38\linewidth}
    \centering
    \includegraphics[width=0.95\textwidth]{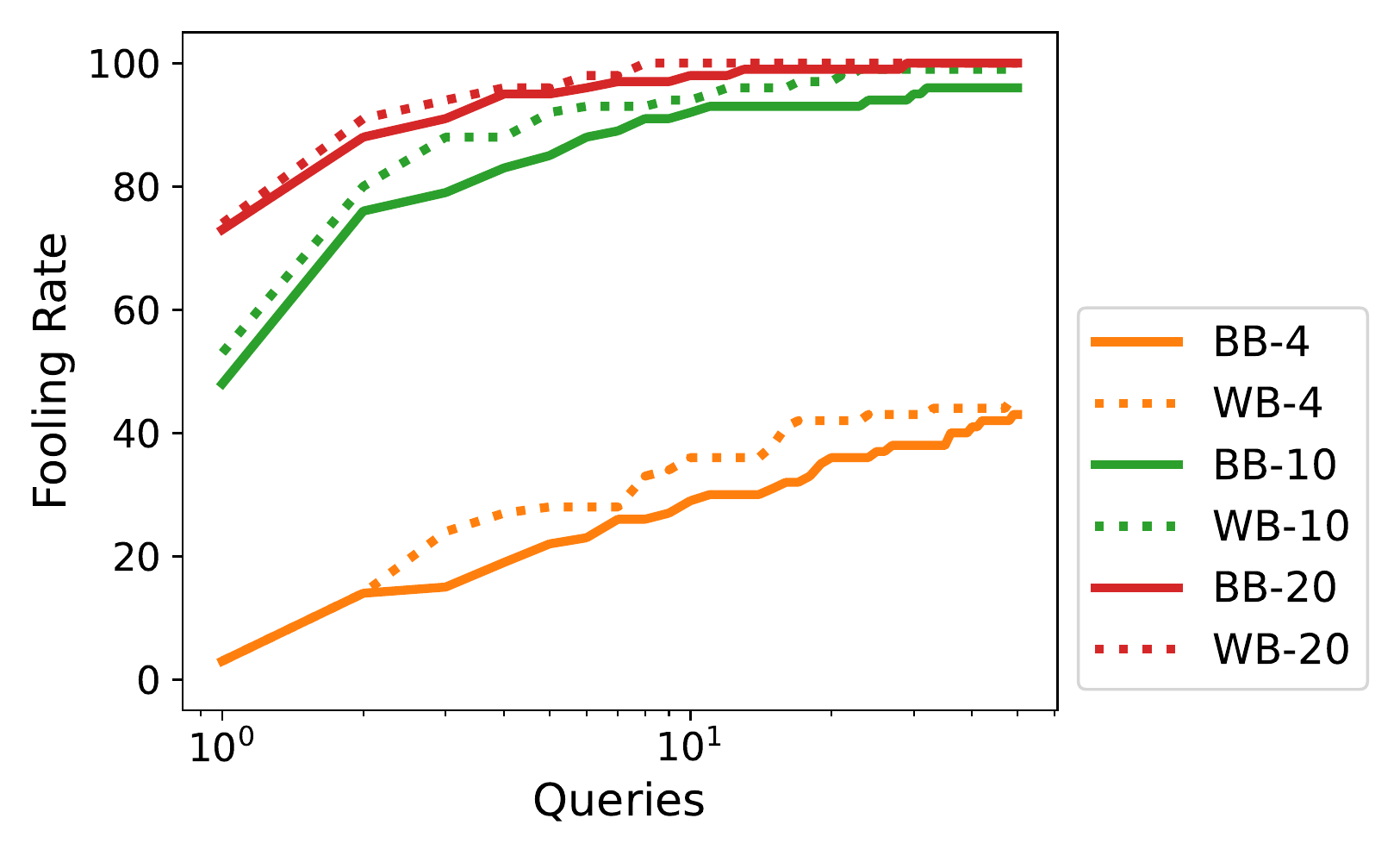}
    \caption{ResNext-50}
\end{subfigure}
\caption{Comparison of targeted attack fooling rate with different number of ensemble models $N \in \{4,10,20\}$ in PM. Every experiment is performed with whitebox gradient (denoted as `WB' with dotted lines) and blackbox score-based coordinate descent (denoted as `BB' with solid lines). Experiment was run on $100$ images.}
\label{fig:compare-wb-bb}
\end{figure}

\textbf{Surrogate ensemble size $(N)$.}
To evaluate the effect of surrogate ensemble size on the performance of our method, we performed targeted blackbox attacks experiment on three different victim models using three different sizes of the surrogate ensemble: $N\in \{4,10,20\}$. The results are presented in Figure \ref{fig:compare-wb-bb} in terms of fooling success rate vs number of queries. 
As we increase the ensemble size, the fooling rate also increases. With $N =20$, the targeted attack fooling rate is almost perfect within $50$ queries. Specifically, for \texttt{VGG-19} with $N=20$, we improve from $54\%$ success rate at the first query (with equal ensemble weights) to $96\%$ success rate at the end of $50$ queries; this equates to $78\%$ improvement. \texttt{DenseNet-121} and \texttt{ResNext-50} can achieve $100\%$ fooling rate with $N=20$. With \texttt{DenseNet-121}, using $10$ surrogate models, we can achieve a fooling rate of $98\%$. While using $4$ models is challenging with respect to all victim models, we can see a rapid and significant improvement in fooling rates when the number of queries increases. 

\textbf{Comparison of whitebox (gradient) vs blackbox (queries).}
To check the effectiveness of our query-based coordinate descent approach for updating $\vw$, we compare its performance with the alternative approach of calculating the exact gradient of victim loss under the whitebox setting. The results are presented in Figure \ref{fig:compare-wb-bb} as dotted lines. We observe that our blackbox query approach provides similar results as the whitebox version, which implies the coordinate-wise update of $\vw$ is as good as a complete gradient update.

\subsection{Hard-label attacks}
The queries generated by our PM are highly transferable and can be used to craft successful attacks for hard-label classifiers. To generate a sequence of queries for hard-label classifiers, we pick a `surrogate victim' model and generate queries by updating $\vw$ in the same manner as the score-based attacks for $Q$ iterations (without termination). We store the queries generated at every iterations in a query set $\{\delta^1,\ldots, \delta^Q\}$. 
We test the victim hard-label blackbox model using $x+\delta$ by selecting $\delta$ from the set in a sequential order until either the attack succeeds or the queries finish. 

\begin{figure}[t]
\centering
\begin{subfigure}[c]{0.325\linewidth}
    \centering
    \includegraphics[width=0.95\textwidth]{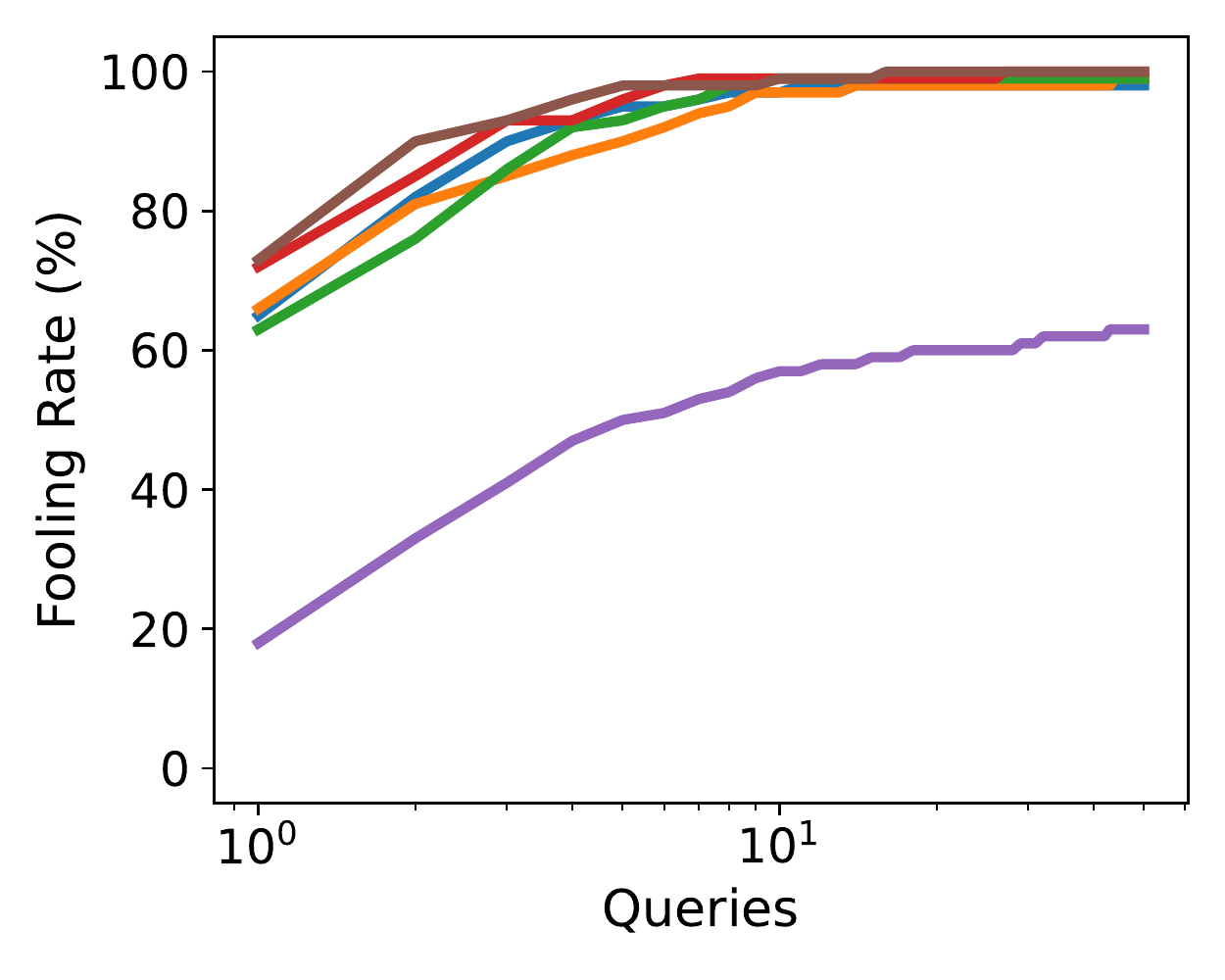}
    \caption{Targeted}
    \label{fig:hard-label-targeted}
\end{subfigure}
\begin{subfigure}[c]{0.45\linewidth}
    \centering
    \includegraphics[width=0.95\textwidth]{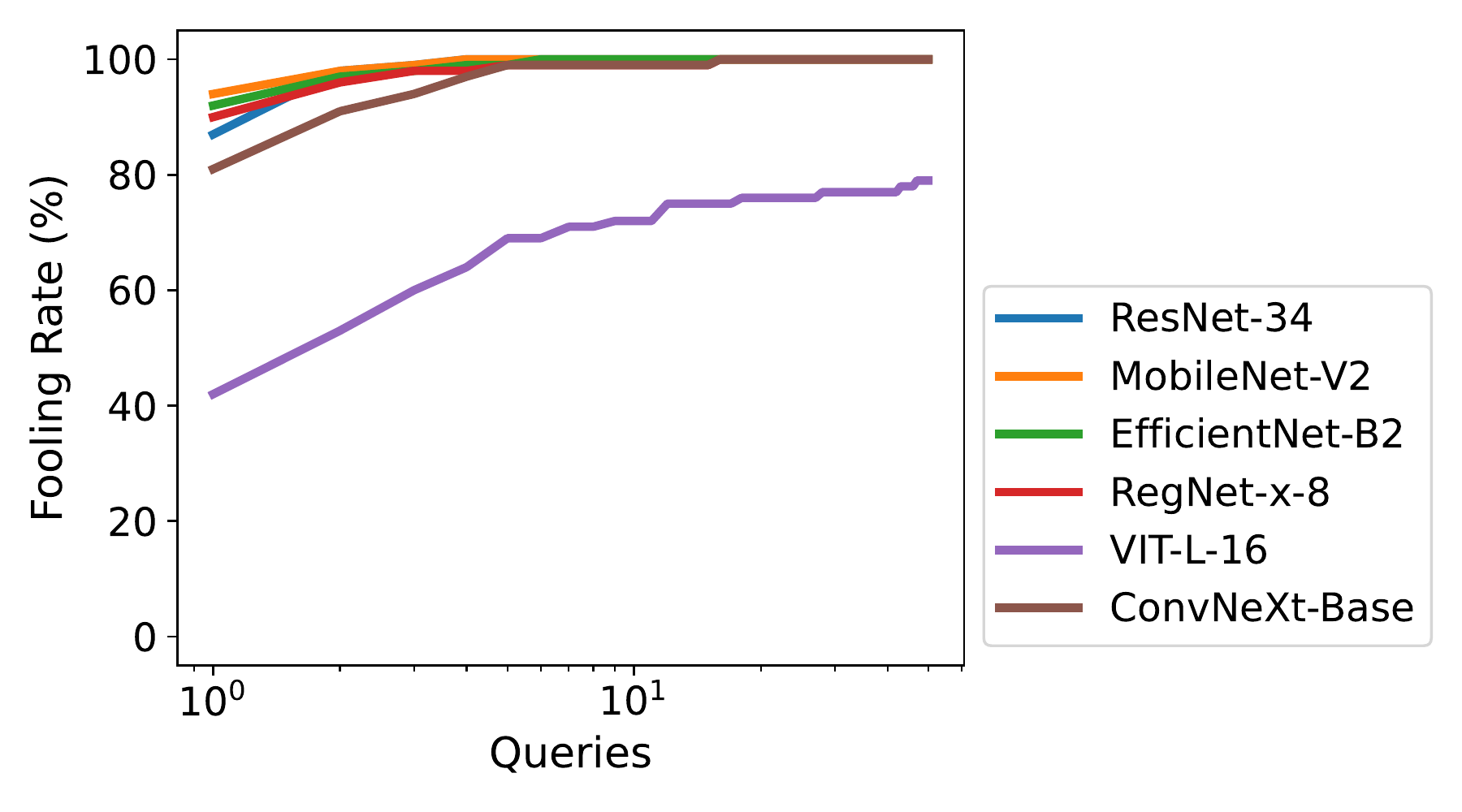}
    \caption{Untargeted}
    \label{fig:hard-label-untargeted}
\end{subfigure}
\caption{Performance of blackbox attack on 6 hard-label classifiers. Our method generates a sequence of queries for targeted attack using \texttt{VGG-19} as a victim model while the PM has $N = 20$ models in the surrogate ensemble. Experiment performed on $100$ images.}
\label{fig:hard-label}
\end{figure}

In our experiments, we observed that this approach can  achieve a high targeted attack fooling rate on a variety of models. 
We present the results of our experiment in Figure \ref{fig:hard-label}, where we report attack success rate vs query count for 6 models: \{\texttt{MobileNet-V2, ResNet-34, ConvNeXt-Base, EfficientNet-B2, RegNet-x-8, VIT-L-16}\}. We used \texttt{VGG-19} as the `surrogate victim' model to generate the queries using the PM with $20$ surrogate models. Using the saved surrogate perturbations, we can fool all models almost 100\%, except for \texttt{VIT-L-16} \cite{dosovitskiy2020image} that is a vision transformer and architecturally very different from the majority of surrogate ensemble models (thus difficult to attack). Nevertheless, the fooling rate increases from $18\% \rightarrow 63\%$, which is a $250\%$ improvement. 

\subsection{Attack on commercial Google Cloud Vision API}
We demonstrate the effectiveness of our approach under a practical blackbox setting by attacking the Google Cloud Vision (GCV) label detection API. GCV detects and extracts information about entities in an image, across a very broad group of categories containing general objects, locations, activities, animal species, and products. Thus, the label set is very different from that of ImageNet, and largely unknown to us. We have no knowledge about the detection models in this API either. We randomly select 100 images from the aforementioned ImageNet dataset that are correctly classified by GCV, and perform untargeted attacks against GCV using $20$ surrogate models with perturbation budget of $\ell_\infty \leq 12$ to align with the setting in TREMBA \cite{huang2019black}.

For each input image, GCV returns a list of labels, which are usually the top 10 labels ranked by probability. Under the success metric of changing the top 1 label to any other label, same as in \cite{huang2019black}, our attack can achieve a fooling rate of $91\%$ with only $2.9$ queries per image on average, which is much lower than $8$ queries TREMBA reported for similar experiment. We present some successful examples in Figure~\ref{fig:gcv-api}.
We present additional results in the supplementary material that show our attacks from classification can transfer to object detection models.

\begin{figure}[h]
\centering
\begin{subfigure}[c]{0.425\linewidth}
    \centering
    \includegraphics[width=1\textwidth]{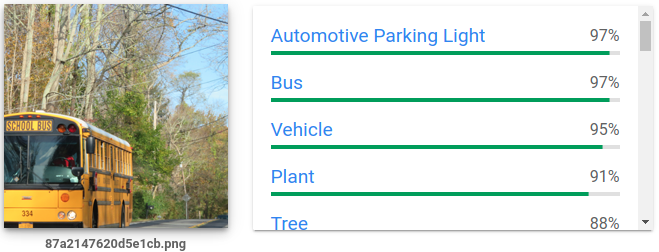}
    \caption{Original Image - Bus}
\end{subfigure}
~~
\begin{subfigure}[c]{0.425\linewidth}
    \centering
    \includegraphics[width=1\textwidth]{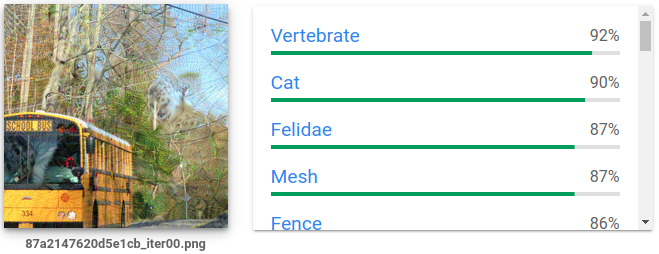}
    \caption{Attacked Image}
\end{subfigure} \\
\vspace{7pt} 
\begin{subfigure}[c]{0.425\linewidth}
    \centering
    \includegraphics[width=1\textwidth]{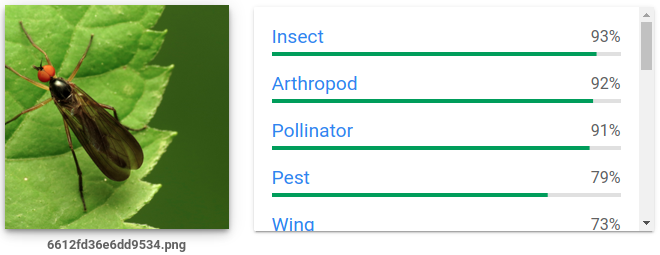}
    \caption{Original Image - Fly}
\end{subfigure}
~~
\begin{subfigure}[c]{0.425\linewidth}
    \centering
    \includegraphics[width=1\textwidth]{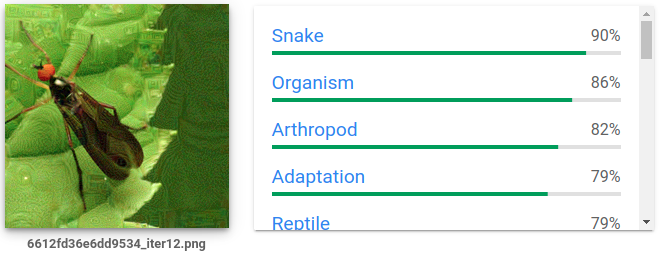}
    \caption{Attacked Image}
\end{subfigure}
\caption{Visualization of some successful attacks on Google Cloud Vision.}
\label{fig:gcv-api}
\end{figure}

\section{Conclusion and discussion}
We propose a novel and simple approach, BASES, to effectively perform blackbox attacks in a query-efficient manner, by searching over the weight space of ensemble models. Our extensive experiments demonstrate that a wide range of models are vulnerable to our attacks at the fooling rate of over 90\% with as few as 3 queries for targeted attacks. The attacks generated by our method are highly transferable and can also be used to attack hard-label classifiers. Attacks on Google Cloud Vision API further demonstrates that our attacks are generalizable beyond the surrogate and victim models in our experiments. 

\textbf{Limitations.} 
1) Our method needs a diverse ensemble for attacks to be successful. Even though the search space is low-dimensional, the generated queries should span a large space so that they can fool any given victim model. 
This is not a major limitation for image classification task as a large number of models are available, but it can be a limitation for other tasks. 
2) Our method relies on the PM to generate a perturbation query for every given set of weights. The perturbation generation over surrogate ensemble is computationally expensive, especially as the ensemble size becomes large. In our experiments, one query generation with $\{4,10,20\}$ surrogate models requires nearly $\{2.4s,9.6s,18s\}$ per image on Nvidia GeForce RTX 2080 TI. Since our method requires a small number of queries, the overall computation time of our method remains small.

\textbf{Societal impacts.} 
We propose an effective and query efficient approach for blackbox attacks. Such adversarial attacks can potentially be used for malicious purposes. Our work can help further explain the vulnerabilities of DNN models and reduce technological surprise. We also hope this work will motivate the community to develop more robust and reliable models, since DNNs are widely used in real-life or even safety-critical applications.

\noindent\textbf{Acknowledgments.} 
This material is based upon work supported by the Defense Advanced Research Projects Agency (DARPA) under agreement number HR00112090096.
Approved for public release; distribution is unlimited.

\small
\bibliographystyle{unsrtnat}
\bibliography{refs}

\newpage

\newpage
\clearpage

\label{sec:appendix}
\noindent \begin{center} {\large  \textbf{Blackbox Attacks via Surrogate Ensemble Search \\ Supplementary Material}} \end{center}

\makeatletter
\def\mysequence#1{\expandafter\@mysequence\csname c@#1\endcsname}
\def\@mysequence#1{%
  \ifcase#1\or -\or - \or -\or -\or -\or A\or B\or C\or D\or E\or F\or G\or H\else\@ctrerr\fi}
\makeatother
\renewcommand\thesection{\mysequence{section}}

\section*{Summary}

In this supplementary material, we provide additional discussion on the selection of hyper parameters, results for classification and object detection tasks, and influence of ensemble weights on the loss landscape of the victim model. Below is a summary of main sections of the supplementary material. 
\begin{enumerate}[label=\Alph*.,leftmargin=8mm]
    \item As mentioned in Section \ref{sec:method} of the main text, we provide experimental results justifying our selection of hyper-parameters, such as the choice of loss function for individual surrogate models, ensemble loss function, step size of PGD attack in PM, and the selection of models in the surrogate ensemble. We select the hyper-parameters that achieve the best performance for our experiments. We also discuss the effects of using different target labels. 
    \item As promised in Section \ref{sec:experiments} of the main text, we provide more details about the comparison with TREMBA and GFCS. We present comparisons with the Simulator Attack \cite{ma2021simulating} and a hybrid attack using query-based square attack \cite{suya2019hybrid, andriushchenko2020square}. We also provide additional comparisons with state-of-the-art methods for untargeted attacks and attacks under $\ell_2$ norm constraints. 
    \item We present experiments and results for vanishing attacks on object detectors. Our results indicate that the proposed approach is also effective for tasks beyond classification.
    \item We present some examples of adversarial images generated in our experiments. 
    \item We analyze the effect of ensemble weights on the loss landscape of different victim models.
    \item We present average top-1 classification accuracy of all the models on clean images. 
\end{enumerate}

\section{Analysis of hyper-parameters}\label{sec:hyperparameters}
\subsection{Hyper-parameters for inner optimization (PM)}
Hyper-parameters can greatly impact the attack performance but can get overlooked sometimes. In our experiments, we analyzed how different hyper-parameters influence the performance of our algorithm. The experiment setup is similar to Figure~\ref{fig:compare-sota-linf-targeted-vgg19} (in main text) using only the first 100 images to speed up experiments (because we observed similar trends using all 1000 images).
\begin{figure}[h]
\centering
\begin{subfigure}[c]{0.325\linewidth}
    \centering
    \includegraphics[width=0.95\textwidth]{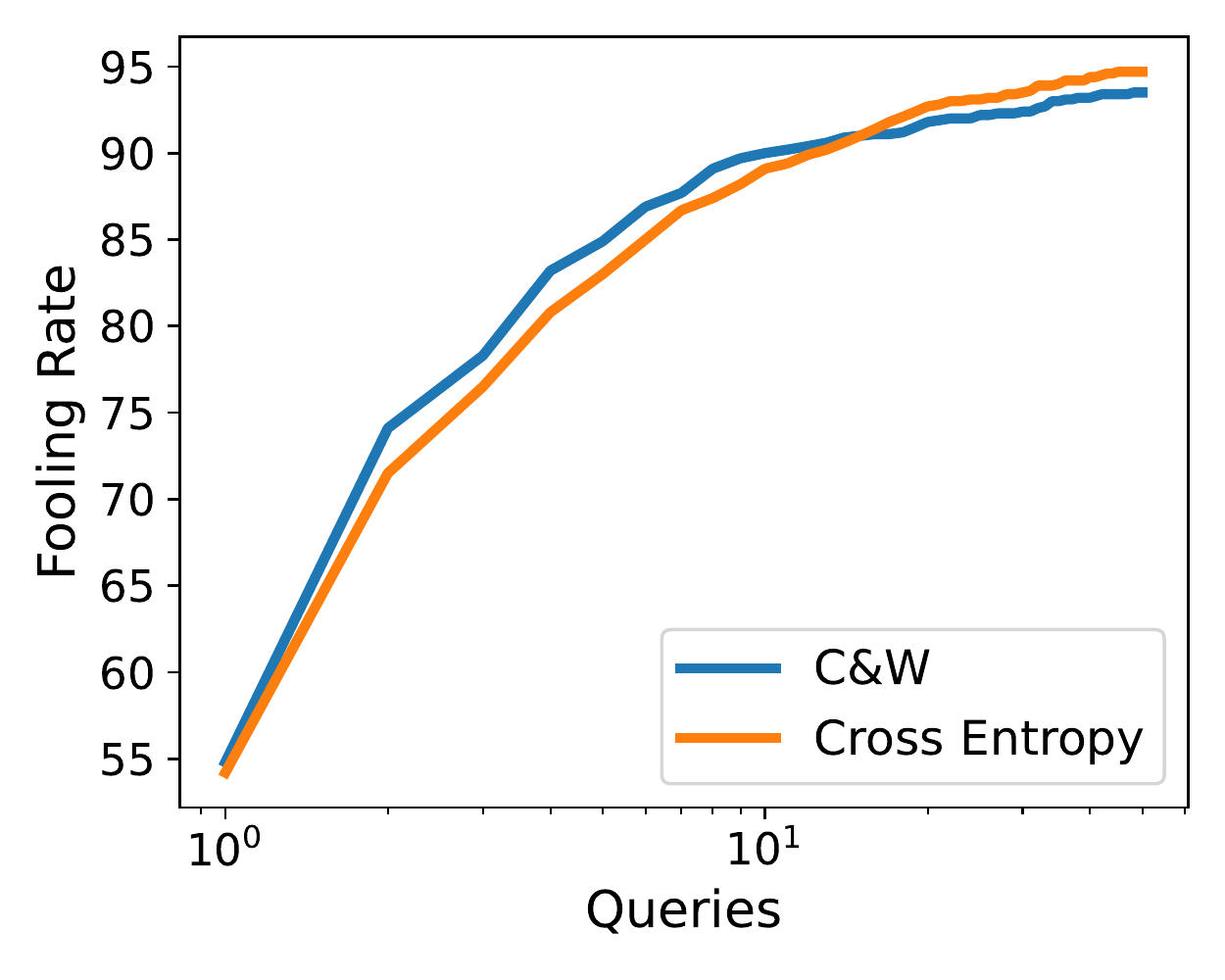}
    \caption{Loss function}
    \label{fig:analysis-loss}
\end{subfigure}
\begin{subfigure}[c]{0.325\linewidth}
    \centering
    \includegraphics[width=0.95\textwidth]{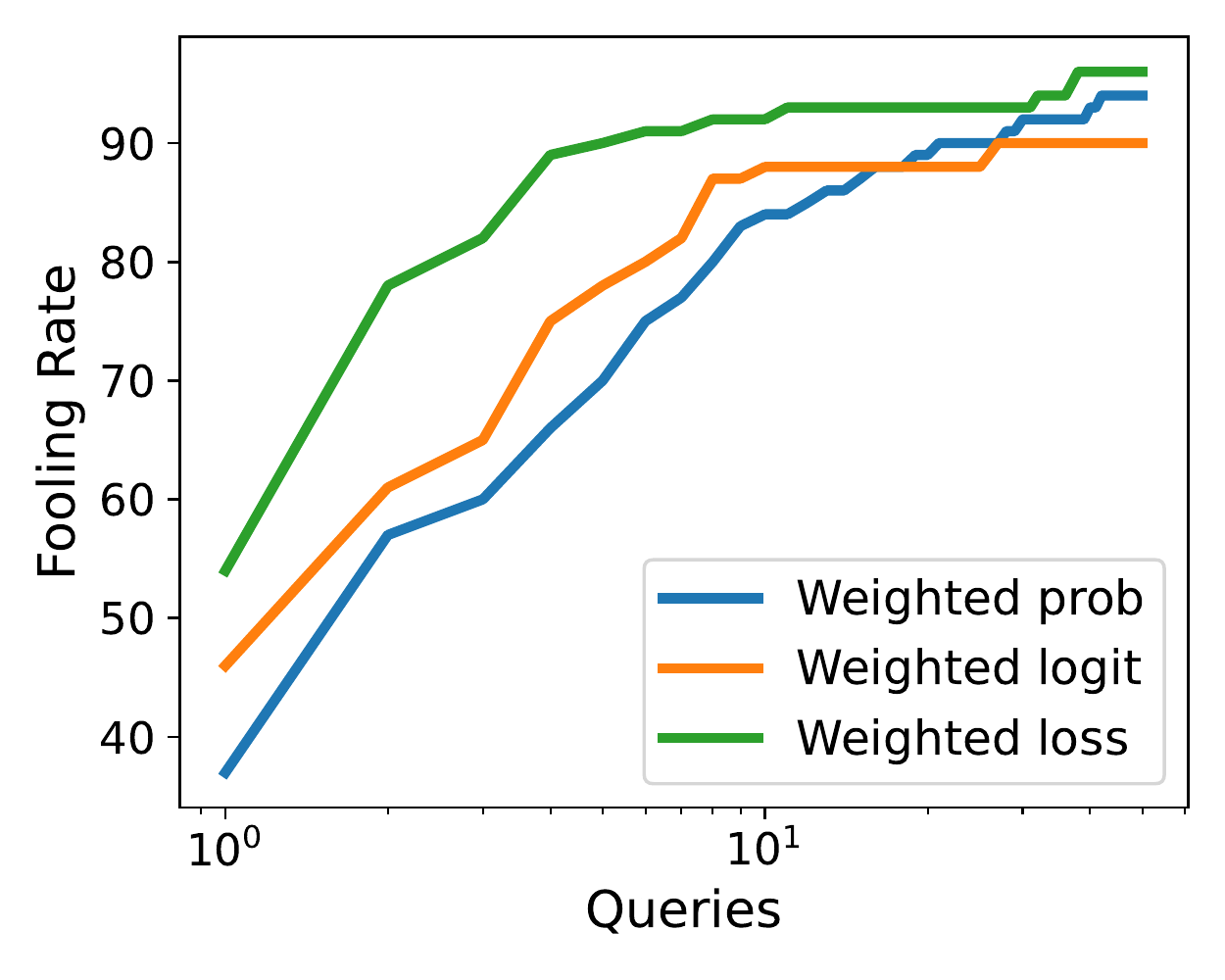}
    \caption{Ensemble loss}
    \label{fig:analysis-fuse}
\end{subfigure}
\begin{subfigure}[c]{0.325\linewidth}
    \centering
    \includegraphics[width=0.95\textwidth]{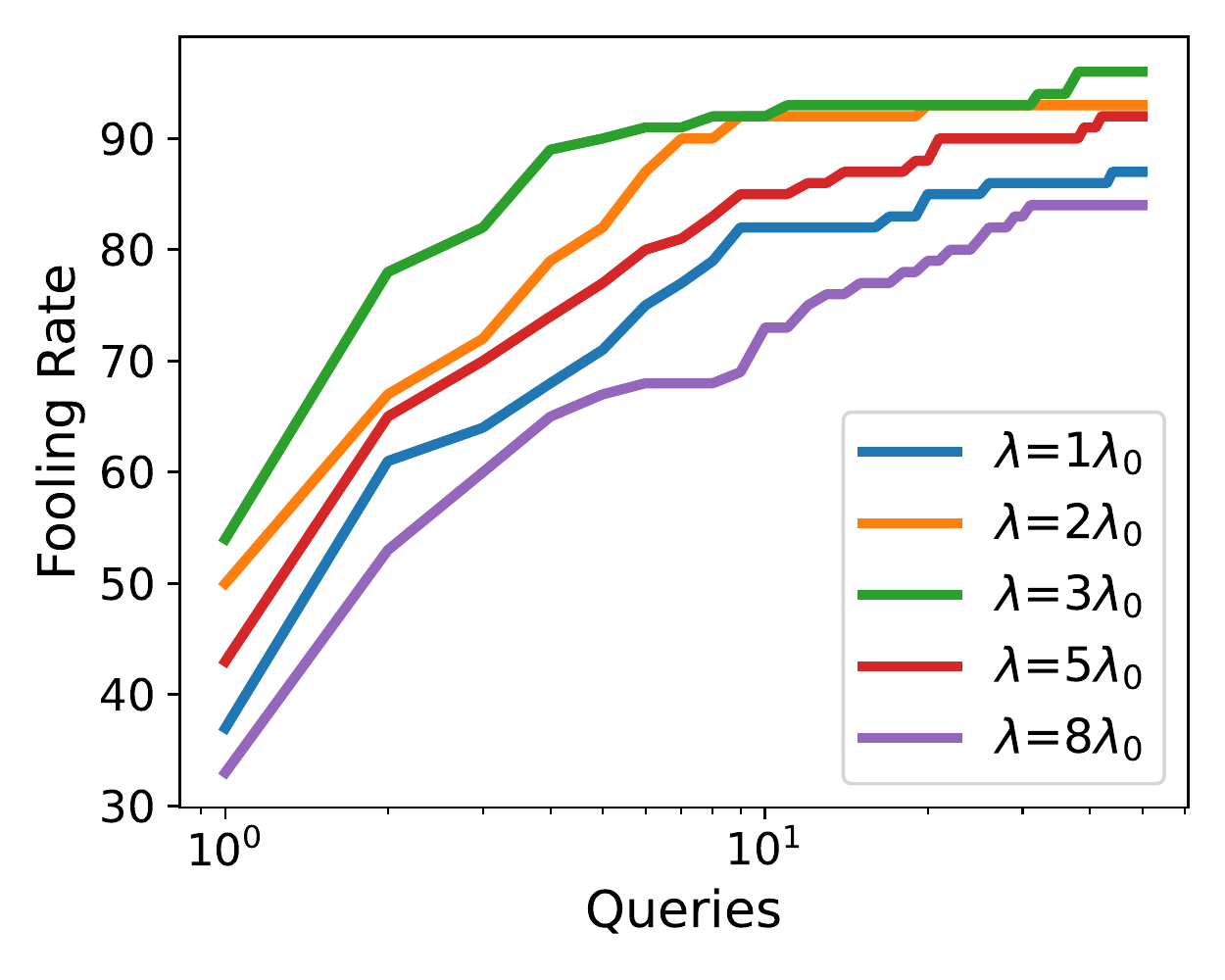}
    \caption{Step size $\lambda$}
    \label{fig:analysis-lambda}
\end{subfigure} \\
\vspace{0.4cm}
\begin{subfigure}[c]{0.325\linewidth}
    \centering
    \includegraphics[width=0.95\textwidth]{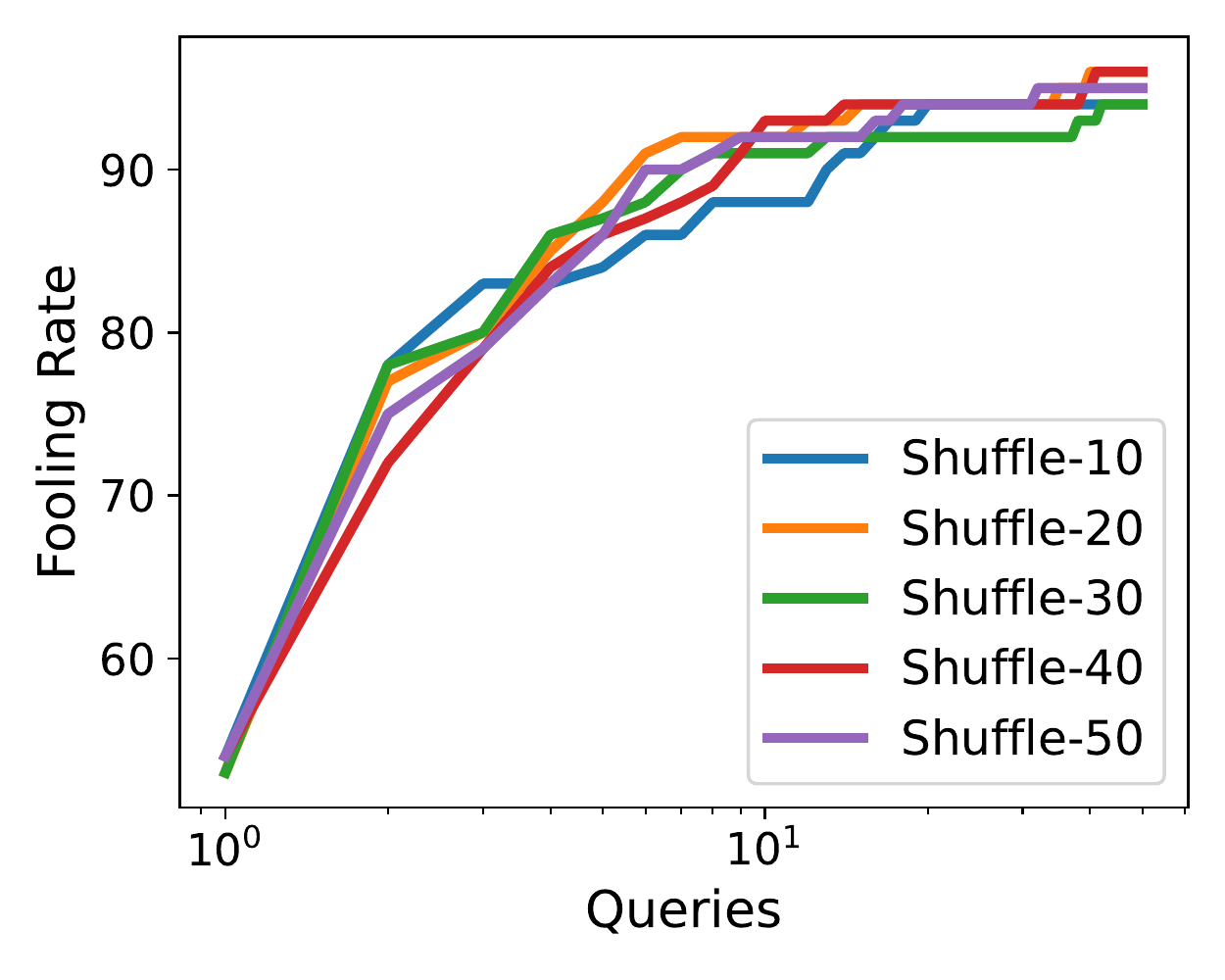}
    \caption{Shuffle model orders}
    \label{fig:analysis-shuffle}
\end{subfigure}
\begin{subfigure}[c]{0.325\linewidth}
    \centering
    \includegraphics[width=0.95\textwidth]{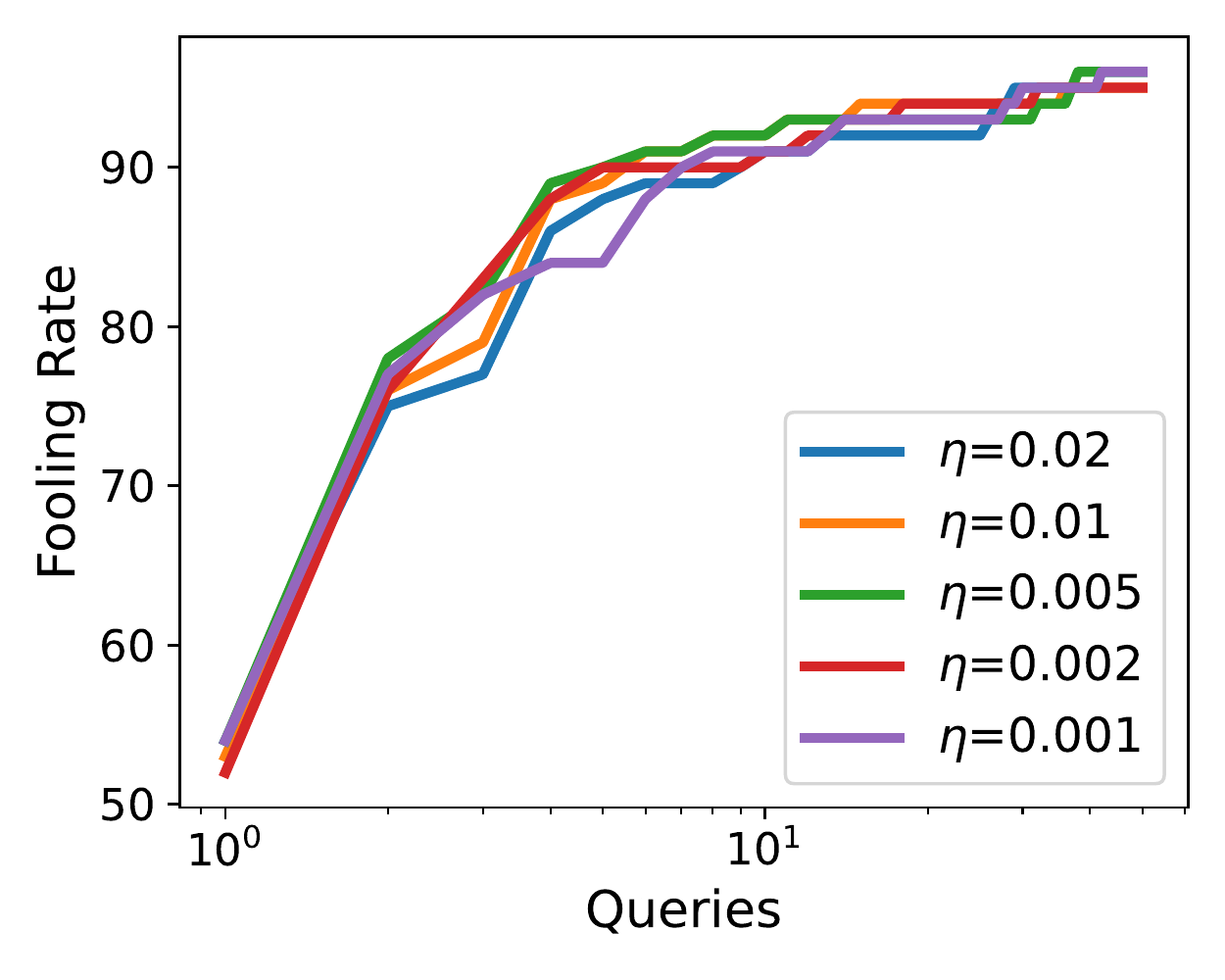}
    \caption{Learning rate $\eta$}
    \label{fig:analysis-lr}
\end{subfigure}
\caption{Analysis on the effect of some hyper-parameters. (a) Loss function for individual surrogate models. (b) Ensemble loss function using three types of weighted combinations. (c) Step size $\lambda$ of inner optimization in PM. (d) Order of surrogate models in PM. (e) Learning rate $\eta$ in updating ensemble weights $\vw$ for outer optimization. All the experiment are performed for targeted attacks on victim model \texttt{VGG-19}, with ensemble size $N = 20$, and evaluated on $100$ images.}
\label{fig:analysis-hyperparameters}
\end{figure}

\textbf{Loss function}. Two popular candidates of loss functions (C\&W, Cross Entropy) show similar performance as shown in Figure \ref{fig:analysis-loss}. We choose C\&W for the sake of convenience in determining success from its sign.

\textbf{Ensemble loss function}. Some previous papers (e.g., MIM \cite{dong2018boosting}) claimed that ensemble with weighted logits (equation \eqref{eq:fuse-logit} in main text) outperforms ensemble with weighted probabilities and weighted combination of loss (equations \eqref{eq:fuse-prob} and \eqref{eq:fuse-loss} in main text). In our experiments, shown in Figure \ref{fig:analysis-fuse}, we observe that weighted combination of surrogate loss functions provide similar or even higher fooling rate compared to weighted probabilities or logits. 

\textbf{PGD step size $\lambda$}. Since we are running the PGD-based attack in PM, the step size $\lambda$ can influence the attack success rate. For perturbation budget $\varepsilon = 16$ and $T = 10$ iterations, I-FGSM will use a step size of $\lambda_0 = \nicefrac{\varepsilon}{T} = 1.6$ to prevent the perturbation from exceeding the $\ell_\infty$ norm bound. Since PGD projects the perturbation back to its feasible set at each iteration, we can increase the step size $\lambda$ such that more pixels saturate, which leads to a higher attack success rate. In Figure \ref{fig:analysis-lambda}, we report results using $\lambda$ as a multiple of $\lambda_0$ using multiplying factors $\{2,3,5,8\}$. We choose the best step size for our experiments, which is $\lambda = 3\lambda_0$.

\textbf{Selection of surrogate models.}
As specified in the main text, our method needs a diverse ensemble for attacks to be successful. Following the setting in TREMBA \cite{huang2019black}, we start with four surrogate models, \texttt{\{VGG-16-BN, ResNet-18, SqueezeNet-1.1, GoogleNet\}}. To improve the diversity of our ensemble, we insert more models from different families. We expand it to ten by adding  \texttt{\{MNASNet-1.0, DenseNet-161, EfficientNet-B0, RegNet-y-400, ResNeXt-101, Convnext-Small\}}. Finally, we add \texttt{\{VGG-13, ResNet-50, DenseNet-201, Inception-v3, ShuffleNet-1.0, MobileNet-v3-Small, Wide-ResNet-50, EfficientNet-B4, RegNet-x-400, VIT-B-16\}} to create an ensemble with $20$ models. We observe that by using $20$ models in the ensemble, our method can already achieve an almost perfect targeted attack fooling rate. 

\subsection{Hyper-parameters for outer optimization}
\textbf{Order of surrogate models}. Since we are using a coordinate descent approach, the order of coordinates (i.e., surrogate models in our case) may influence the performance of our method. We performed different experiments by shuffling the order of the models using different random seeds (10, 20, 30, 40, 50). The results in Figure~\ref{fig:analysis-shuffle} suggest that our method provides identical results for different sequences of surrogate models.

\textbf{Learning rate $\eta$}. Learning rate is often an important hyper-parameter that can influence performance, and we selected our learning rate to be $\nicefrac{1}{10}{th}$ of the average ensemble weight with $20$ models (i.e., $\eta = 0.005$). We compare different learning rates in the range of $\{0.02, 0.01, 0.005, 0.002, 0.001\}$ while ensuring that all individual surrogate weights remain  non-negative. The results in Figure~\ref{fig:analysis-lr} suggest that our approach is robust to variations in the learning rates.

\subsection{Selection of target labels}
Different selections of target labels may result in different levels of difficulty in attacks. Here we evaluate different proposals for selecting the target labels, including the `easiest' label (the label with the second highest  original confidence score), the `hardest' label (the label with the lowest original confidence score), and a random label. We performed an experiment to test the difficulty of three types of target labels and report our results in Table~\ref{tab:target-labels} below. Corresponding to Table \ref{tab:pros-cons} in our paper, we use \texttt{DenseNet-121} as the victim model. Here we randomly select 100 images for evaluation. We see that the original confidence scores of the NeurIPS17 target classes are already close to 0 (which means they are already challenging cases), for which our method requires an average of $1.64$ queries to achieve a $100\%$ fooling rate. The second most likely label has an average confidence of $0.08$ (whereas top 1 is $0.8$), and our method achieves a 100\% fooling rate with an average of $1.02$ queries. For the `hardest' setting of the least confident class label, our method shows a slight drop in fooling success rate and achieves $97\%$ success using $2.20$ queries on average.

\begin{table}[h]
\centering
\caption{Performance of BASES on different selection of target labels. }
\label{tab:target-labels}
\small
\begin{tabular}{ccccccc}
\hline
\multirow{2}{*}{Target classes} & \multirow{2}{*}{Avg. confidence} & \multirow{2}{*}{Fooling rate} & \multicolumn{4}{c}{Query counts} \\ \cline{4-7} 
 &  &  & $\mathbf{mean}$ & $\mathbf{min}$ & $\mathbf{max}$ & $\mathbf{median}$ \\ \hline
`Easiest’ & 0.08 & 100\% & 1.02 & 1 & 2 & 1 \\
NeurIPS17 & $8.92\times10^{-6}$ & 100\% & 1.64 & 1 & 13 & 1 \\
`Hardest’ & $1.74\times10^{-8}$ & 97\% & 2.20 & 1 & 15 & 1 \\ \hline
\end{tabular}
\end{table}

\section{Experiments on classification}

\textbf{Comparison with TREMBA.}
TREMBA~\cite{huang2019black} requires one trained generator for each target class; thus, it is not feasible to test it for any arbitrary target label selected from 1000 classes in ImageNet. For a fair comparison, we attack each image using one of the 6 target labels available in trained TREMBA model $\{0,20,40,60,80,100\}$ and average the query counts. Furthermore, TREMBA generator was trained using an ensemble of $4$ surrogate models; while it is possible to train the generator with more surrogate models, training one generate per target label is expensive and non-trivial in terms of hyper-parameter tuning. Therefore, in our experiments, we used the trained generator from the paper.
It is worth pointing out that our method with 4 surrogate models (as shown in Figure~\ref{fig:compare-wb-bb}) is still better than TREMBA in the low query count regime. TREMBA can provide better success rate at the expense of increased queries.  

\textit{Why is our method better than TREMBA?} TREMBA generates patterns by optimizing over the latent code of a trained generator, which contributes to the high success rate. TREMBA generator has a large enough range that it can generate adversarial perturbations that fool a victim model. Our experimental results suggests that the space of perturbations generated by our PM (via weighted surrogate ensemble) is better (in terms of diversity and low dimensionality) than TREMBA's generator. That is the reason why we see a steep slope for the first few queries in our success vs query curve. 

\begin{figure}[h]
\centering
\begin{subfigure}[c]{0.3\linewidth}
    \centering
    \includegraphics[width=0.95\textwidth]{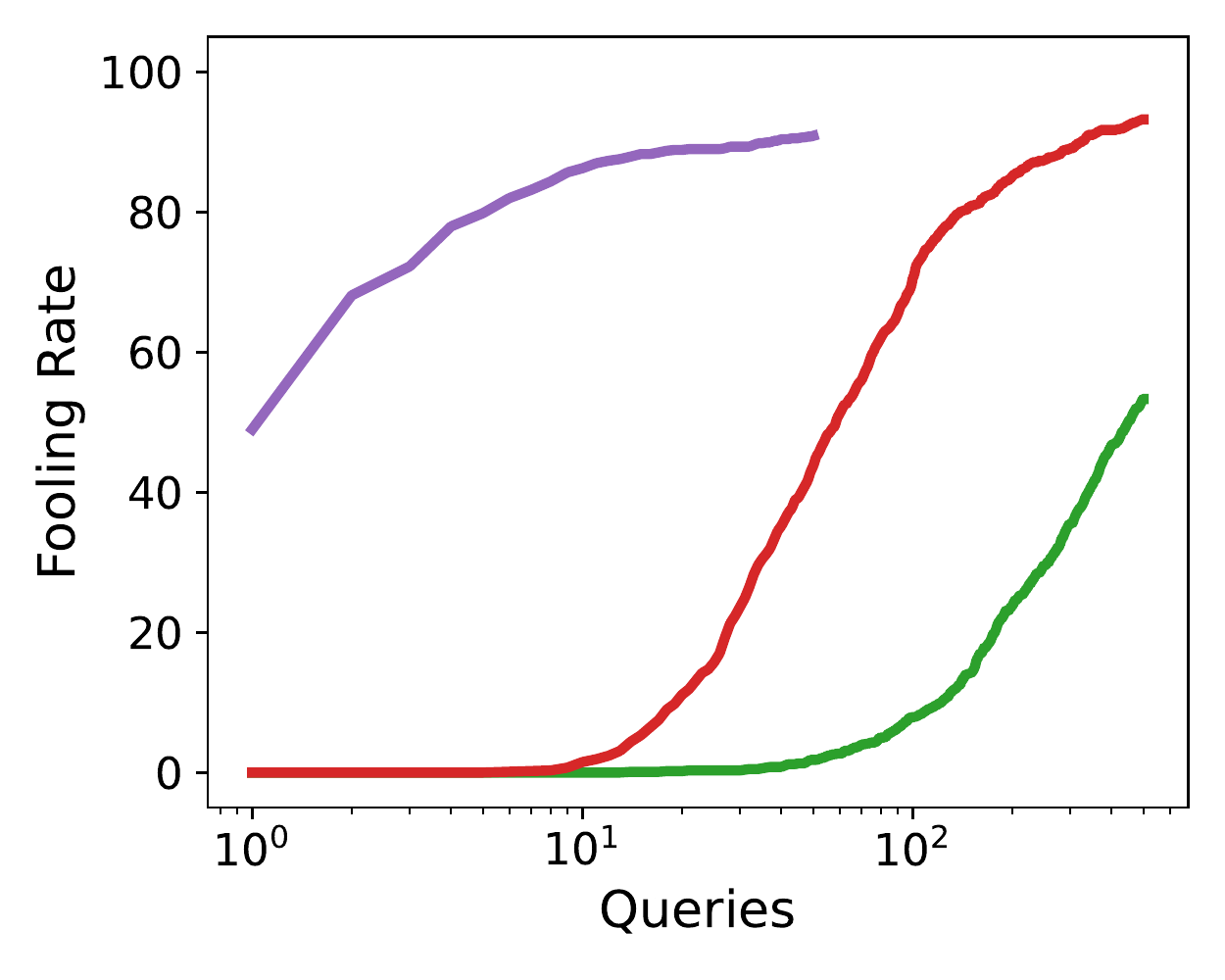}
    \caption{VGG-19}
\end{subfigure}
\begin{subfigure}[c]{0.3\linewidth}
    \centering
    \includegraphics[width=0.95\textwidth]{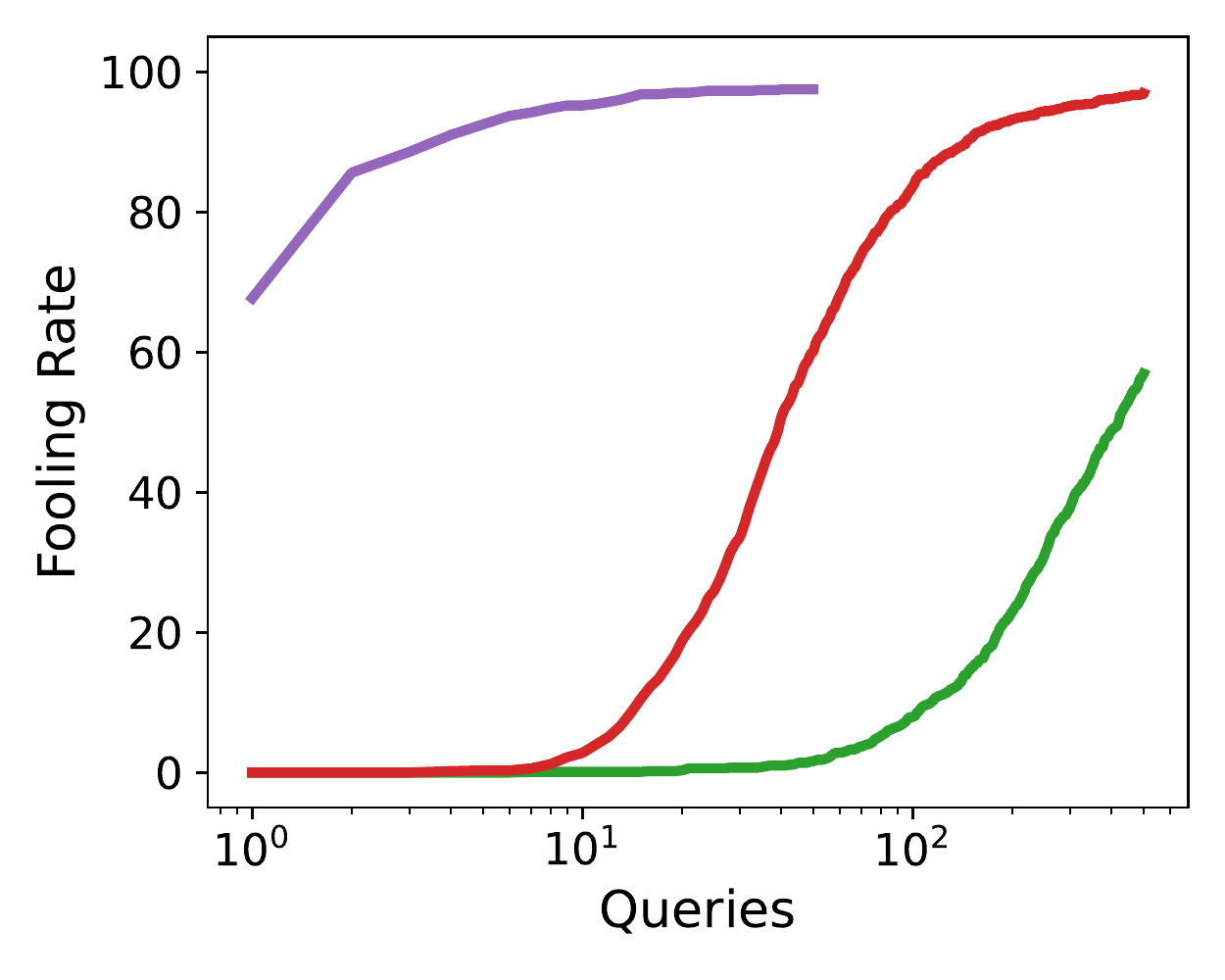}
    \caption{DenseNet-121}
\end{subfigure}
\begin{subfigure}[c]{0.36\linewidth}
    \centering
    \includegraphics[width=0.95\textwidth]{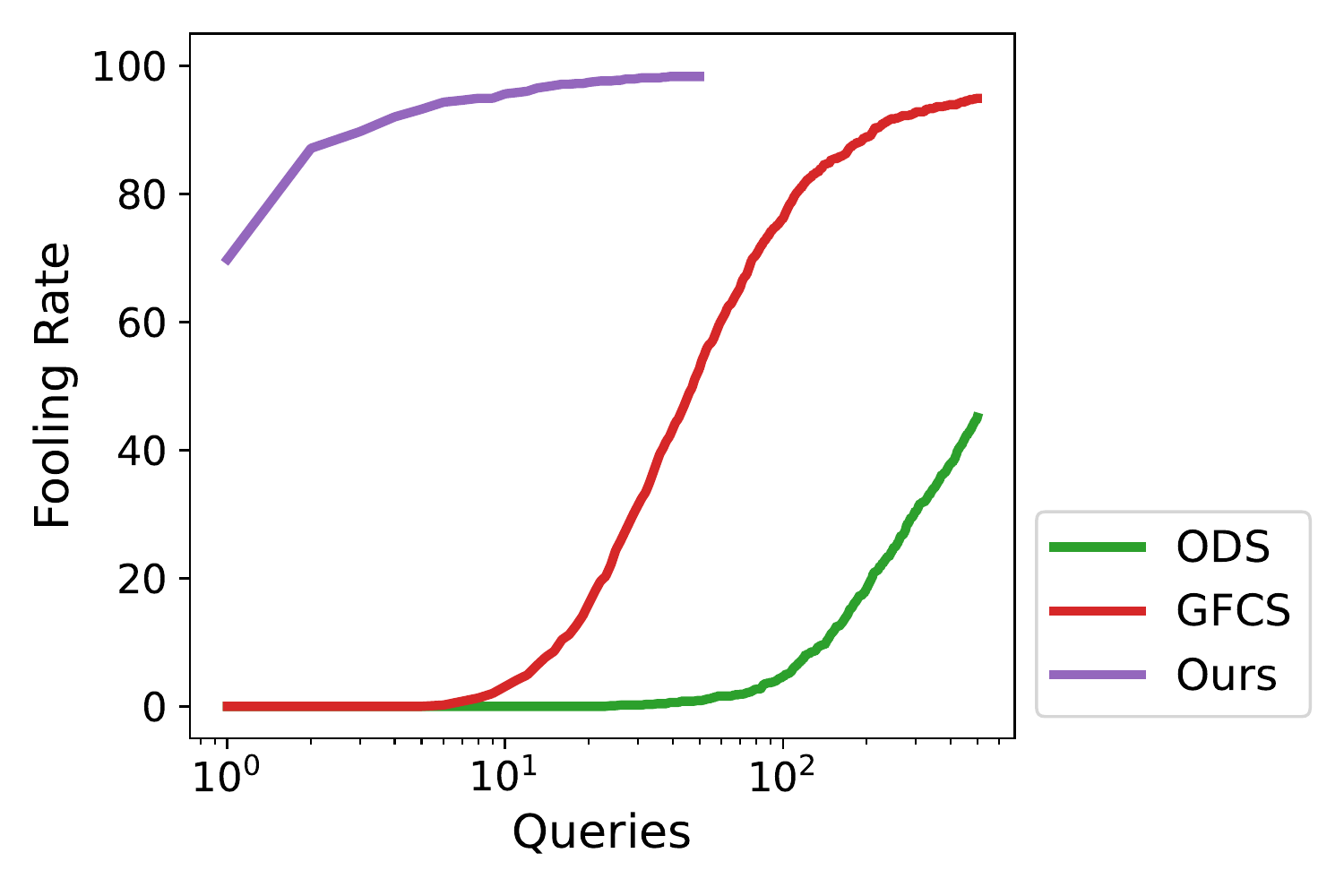}
    \caption{ResNext-50}
\end{subfigure}
\caption{Adversarial attacks generated with $\ell_2$ constraint (equivalent to Figure \ref{fig:compare-sota-linf-targeted} in main text that uses $\ell_\infty$ constraints). Comparison of our method with GFCS / ODS on three victim models under perturbation budget $\ell_2 \leq 3128$ for targeted attacks.}
\label{fig:compare-sota-l2-targeted}
\end{figure}

\textbf{Comparison with GFCS.}
To perform our experiments, we used the same set of $N=20$ surrogate models for GFCS \cite{lord2022attacking} that are used in our PM. GFCS used $\ell_2$ norm constraint and did not compare with TREMBA. While our method can generate perturbations with $\ell_2$ and $\ell_\infty$ constraints, TREMBA generates perturbations with $\ell_\infty$ constraint. To perform a fair comparison, we modified GFCS code to have $\ell_\infty$ constraint and tuned the hyper-parameters to achieve the best performance. 
The step-size is the key parameter that we choose as $0.005$ after searching over a grid of $\{0.2, 0.02, 0.01, 0.005, 0.001, 0.0005\}$. As shown in \ref{fig:compare-sota-l2-targeted}, the performance reported in Figure~\ref{fig:compare-sota-linf-targeted} for $\ell_\infty$ attacks is on par with the performance achieved with original settings of $\ell_2$ norm constraint.

\textit{Why is our method better than GFCS?} Our method is more query efficient because we leverage all surrogate models for each query, whereas GFCS only uses one surrogate model per query. We can see that our method has the steepest slope in Figure~\ref{fig:compare-sota-l2-targeted} and the highest success at the starting point.

\textit{Note about P-RGF}. The original implementation of P-RGF is in Tensorflow, but to unify the platform, we use the Pytorch implementation provided by GFCS \cite{code-gfcs}.

\textbf{Comparison with Simulator Attack.}
We use the same setting as in simulator attack \cite{ma2021simulating} that tests 3 victim blackbox models \texttt{\{DenseNet-121, ResNeXt-101 (32$\times$4d), ResNeXt-101 (64$\times$4d)\}} and uses 16 surrogate models \texttt{\{VGG-11/13/16/19, VGG-11/13/16/19 (BN), ResNet-18/34/50/101/152, DenseNet-161/169/201\}}. All of these models are trained on TinyImageNet \cite{russakovsky2015imagenet} dataset and we obtain the pretrained weights from \cite{ma2021simulating}. We randomly select 1000 tinyImageNet images and use incremental target label selection for targeted attacks. Target label $y_{adv} = (y+1) \mod C$, where $y$ is the original label and total number of classes is $C=200$. Perturbation budget for targeted attack is $\ell_2 \leq 4.6\times255 = 1173$, and for untargeted attack $\ell_\infty \leq 8$. As shown in Table~\ref{tab:simulator}, we achieve perfect fooling rates with less than two queries on average for both targeted and untargeted attacks. Specifically, for \texttt{ResNeXt-101 (32$\times$4d)}, we achieve $100\%$ targeted fooling rate with an average query count of $2.0$, (min = 1, max = 26, median = 1). %
In contrast, simulator attack achieves $84.9\%$ fooling rate using $2558$ queries, which is $1279 \times$ more expensive than ours. For untargeted attack, the trend is similar that our method is $811$--$1445\times$ more query efficient than simulator attack.

\begin{table}[h]
\centering
\small
\caption{Number of queries vs fooling rate of different methods on TinyImageNet dataset.}
\label{tab:simulator}
\begin{tabular}{lcccccc}
\hline
\multirow{3}{*}{Method} & \multicolumn{6}{c}{Number of queries ($\mathbf{mean} / \mathbf{median}$) per image and fooling rate} \\
 & \multicolumn{2}{c}{DenseNet-121} & \multicolumn{2}{c}{ResNeXt-101 (32$\times$4d)} & \multicolumn{2}{c}{ResNeXt-101 (64$\times$4d)} \\ \cline{2-7} 
 & Targeted & Untargeted & \begin{tabular}[c]{@{}c@{}}Targeted\\ \end{tabular} & \begin{tabular}[c]{@{}c@{}}Untargeted\\ \end{tabular} & \begin{tabular}[c]{@{}c@{}}Targeted\\ \end{tabular} & \begin{tabular}[c]{@{}c@{}}Untargeted\\ \end{tabular} \\ \hline
 \vspace{0.1cm}
NES \cite{ilyas2018black} & \begin{tabular}[c]{@{}c@{}}4625 \texttt{/} 4337\\ 88.5\%\end{tabular} & \begin{tabular}[c]{@{}c@{}}1306 / 510\\ 74.3\%\end{tabular} & \begin{tabular}[c]{@{}c@{}}4959 / 4703\\ 88.0\%\end{tabular} & \begin{tabular}[c]{@{}c@{}}2104 / 765\\ 45.3\%\end{tabular} & \begin{tabular}[c]{@{}c@{}}4758 / 4440\\ 88.2\%\end{tabular} & \begin{tabular}[c]{@{}c@{}}2078 / 816\\ 45.5\%\end{tabular} \\
\vspace{0.1cm}
Meta \cite{Du2020Query-efficient} & \begin{tabular}[c]{@{}c@{}}5420 / 5506\\ 24.2\%\end{tabular} & \begin{tabular}[c]{@{}c@{}}3789 / 3202\\ 71.1\%\end{tabular} & \begin{tabular}[c]{@{}c@{}}5440 / 5249\\ 21.0\%\end{tabular} & \begin{tabular}[c]{@{}c@{}}4101 / 3712\\ 33.8\%\end{tabular} & \begin{tabular}[c]{@{}c@{}}5661 / 5250\\ 18.2\%\end{tabular} & \begin{tabular}[c]{@{}c@{}}4012 / 3649\\ 36.0\%\end{tabular} \\
\vspace{0.1cm}
Bandits \cite{ilyas2018prior} & \begin{tabular}[c]{@{}c@{}}2724 / 1860\\ 85.1\%\end{tabular} & \begin{tabular}[c]{@{}c@{}}964 / 520\\ 99.2\%\end{tabular} & \begin{tabular}[c]{@{}c@{}}3550 / 2700\\ 72.2\%\end{tabular} & \begin{tabular}[c]{@{}c@{}}1737 / 954\\ 94.1\%\end{tabular} & \begin{tabular}[c]{@{}c@{}}3542 / 2854\\ 72.4\%\end{tabular} & \begin{tabular}[c]{@{}c@{}}1662 / 1014\\ 95.3\%\end{tabular} \\
\vspace{0.1cm}
Simulator \cite{ma2021simulating} & \begin{tabular}[c]{@{}c@{}}1959 / 1399\\ 89.8\%\end{tabular} & \begin{tabular}[c]{@{}c@{}}811 / 431\\ 99.4\%\end{tabular} & \begin{tabular}[c]{@{}c@{}}2558 / 1966\\ 84.9\%\end{tabular} & \begin{tabular}[c]{@{}c@{}}1380 / 850\\ 96.8\%\end{tabular} & \begin{tabular}[c]{@{}c@{}}2488 / 1982\\ 83.9\%\end{tabular} & \begin{tabular}[c]{@{}c@{}}1445 / 878\\ 97.9\%\end{tabular} \\
\textbf{Ours} & \textbf{\begin{tabular}[c]{@{}c@{}}1.5 / 1\\ 100.0\%\end{tabular}} & \textbf{\begin{tabular}[c]{@{}c@{}}1.0 / 1\\ 100.0\%\end{tabular}} & \textbf{\begin{tabular}[c]{@{}c@{}}2.0 / 1\\ 100.0\%\end{tabular}} & \textbf{\begin{tabular}[c]{@{}c@{}}1.0 / 1\\ 100.0\%\end{tabular}} & \textbf{\begin{tabular}[c]{@{}c@{}}2.0 / 1\\ 100.0\%\end{tabular}} & \textbf{\begin{tabular}[c]{@{}c@{}}1.0 / 1\\ 100.0\%\end{tabular}} \\ \hline
\end{tabular}
\end{table}

\textbf{Comparison with combining transfer and query-based attacks.}
Hybrid attack in \cite{suya2019hybrid} is one of the earliest works that combines transfer and query-based attacks. It uses surrogate models to generate the initial query, which is later updated using feedback from the blackbox victim model via pure query-based methods. To verify that our proposed method is advantageous, we use the perturbations generated by our ensemble models with equal weights as the initial query, and for every failed query we deploy a powerful pure query-based method square attack \cite{andriushchenko2020square}. Following the same setting as in our Table \ref{tab:pros-cons} and Figure \ref{fig:compare-sota-linf-targeted}, we perform targeted attack on \texttt{DenseNet-121} with a perturbation budget of $\ell_\infty \leq 16$. The transfer rate of initial perturbed images is $75.5\%$. We attack the remaining $24.5\%$ failed perturbed images using square attack by allowing a maximum query count of $500$ (same setting as other baseline methods). On this subset of images, we observed a $33.9\%$ fooling rate with query count ($\mathbf{mean}\pm\mathbf{std}$): $238.2 \pm 127.4$. Overall, including the images that can initially transfer, the combination of \cite{suya2019hybrid} and \cite{andriushchenko2020square} achieves a fooling rate of $83.8\%$, with query count ($\mathbf{mean}\pm\mathbf{std}$): $24.5 \pm 81.4$. In comparison, our method achieves a 99.4\% fooling rate with a query count of $1.8 \pm 2.7$. Similar trends appear for other victim models, as shown in Table \ref{tab:hybrid-square}. Our main takeaway is that even though the surrogate ensemble provides highly transferable perturbation or perturbations that can be used as initialization for query-based optimization methods. The query-based methods lose their advantage by querying over a high dimensional image space. Our method searches over the weights of the ensemble loss, which is very low dimension and provides query efficiency.

\begin{table}[t]
\centering
\small
\caption{Number of queries vs fooling rate for hybrid methods that combine transfer and query-based attacks.}
\label{tab:hybrid-square}
\begin{tabular}{lcc}
\hline
\multirow{2}{*}{Models} & \multicolumn{2}{c}{Fooling rate and number of queries ($\mathbf{mean}\pm\mathbf{std}$) per image} \\
 & Combine \cite{suya2019hybrid} and \cite{andriushchenko2020square} & \textbf{Ours} \\ \hline
\vspace{0.1cm}
VGG-19 & 64.7\% ; 34.1 $\pm$ 99.2 & \textbf{95.9\% ; 3.0 $\pm$ 5.4} \\
\vspace{0.1cm}
DenseNet-121 & 83.8\% ; 24.5 $\pm$ 81.4 & \textbf{99.4\% ; 1.8 $\pm$ 2.7} \\
ResNext-50 & 84.3\% ; 24.7 $\pm$ 81.4 & \textbf{99.7\% ; 1.8 $\pm$ 2.6} \\ \hline
\end{tabular}
\end{table}

\begin{figure}[h]
\centering
\begin{subfigure}[c]{0.3\linewidth}
    \centering
    \includegraphics[width=0.95\textwidth]{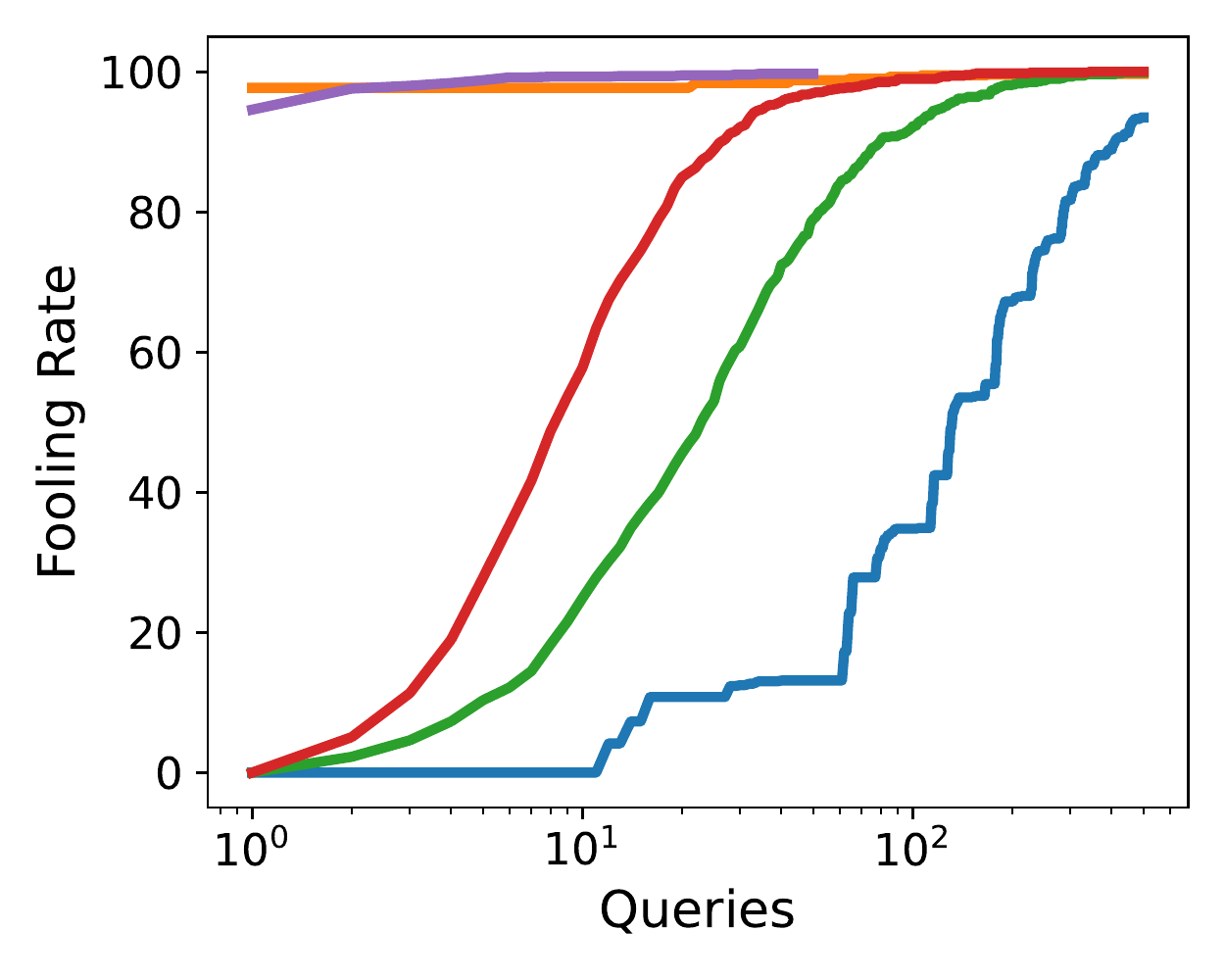}
    \caption{VGG-19}
\end{subfigure}
\begin{subfigure}[c]{0.3\linewidth}
    \centering
    \includegraphics[width=0.95\textwidth]{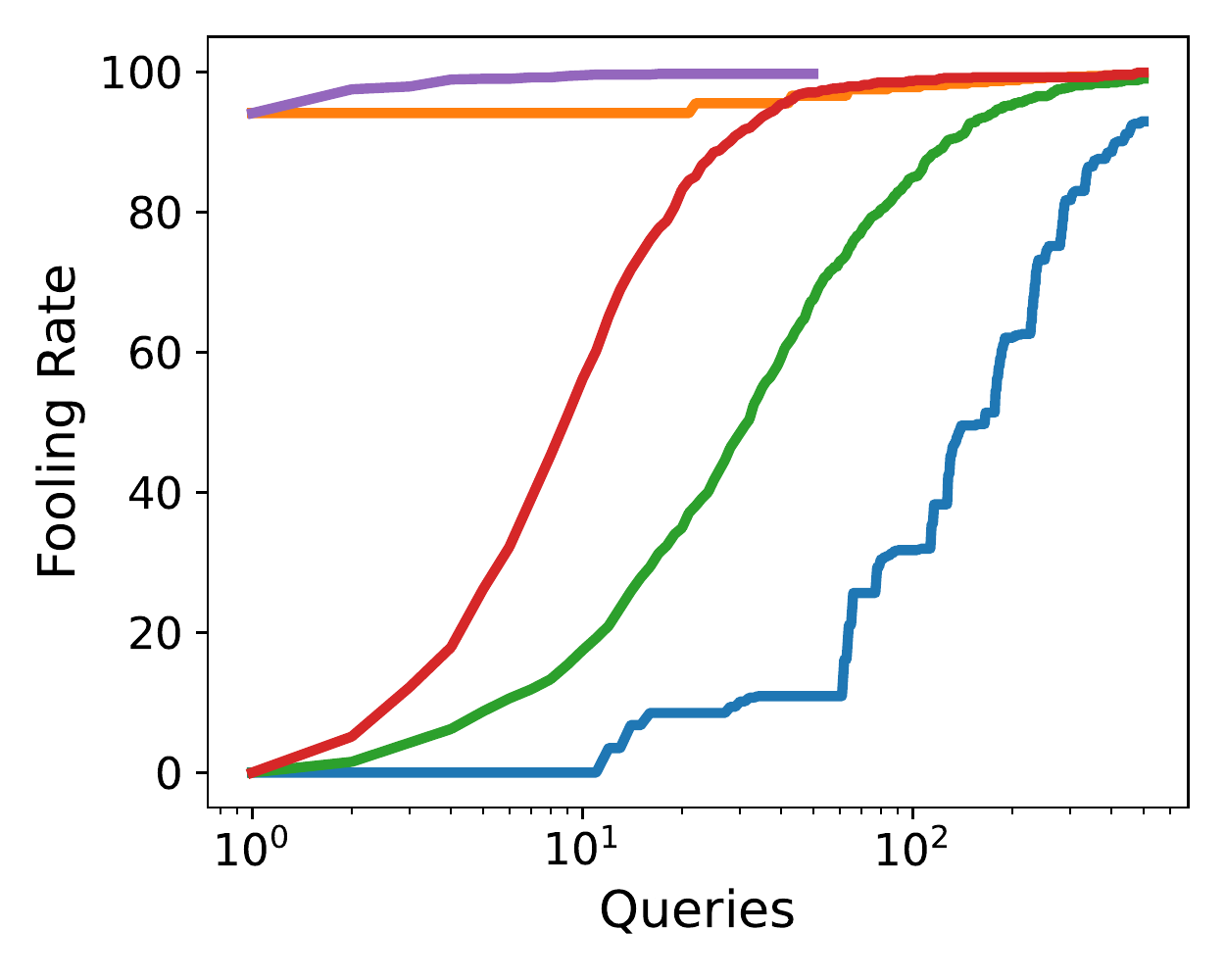}
    \caption{DenseNet-121}
\end{subfigure}
\begin{subfigure}[c]{0.38\linewidth}
    \centering
    \includegraphics[width=0.99\textwidth]{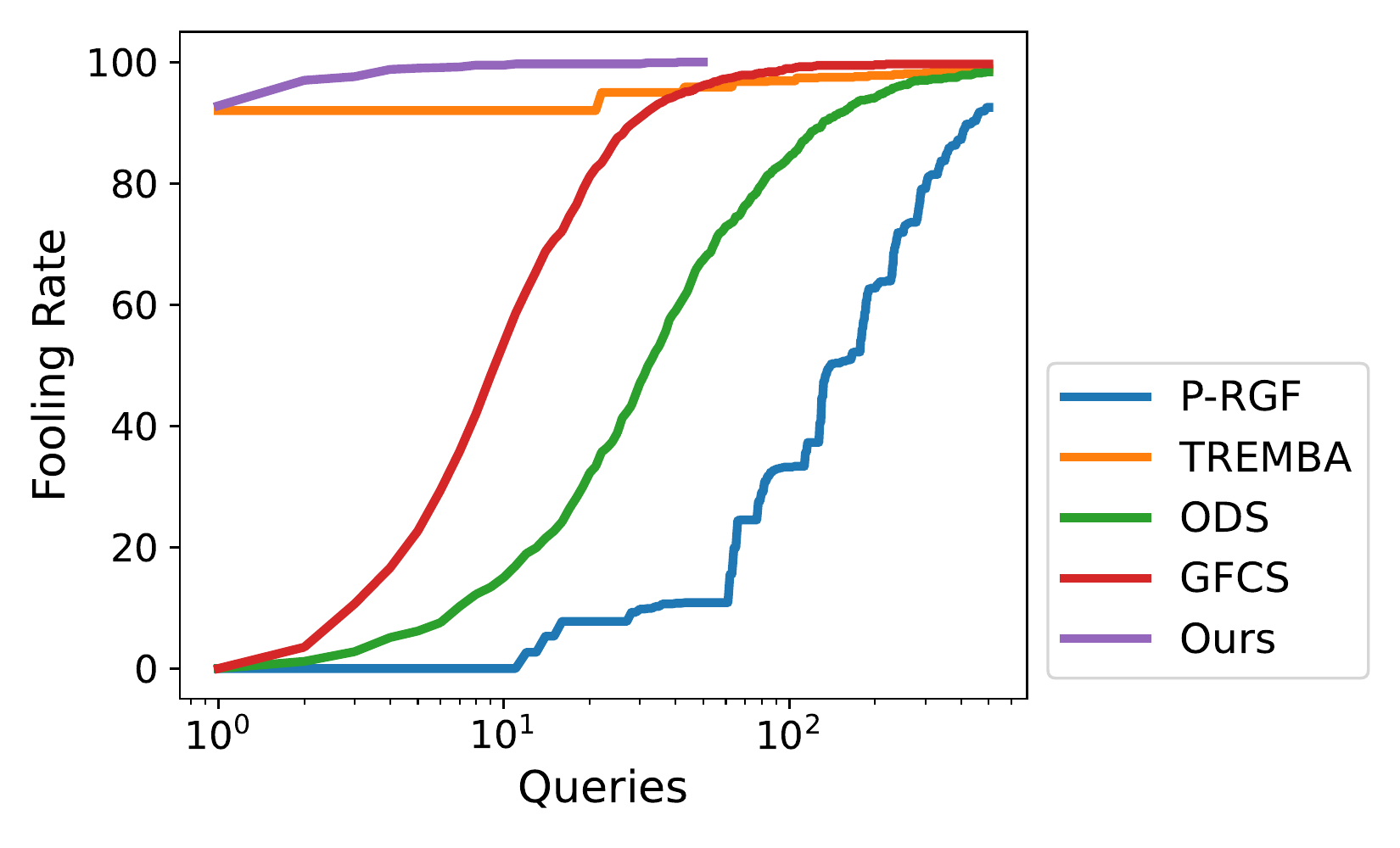}
    \caption{ResNext-50}
\end{subfigure}
\caption{Untargeted attacks (version of Figure \ref{fig:compare-sota-linf-targeted} in the main text). Comparison of 5 attack methods on three victim models under perturbation budget $l_\infty \leq 16$ for untargeted attack. All methods can achieve near perfect success rate within 500 queries.}
\label{fig:compare-sota-linf-untargeted}
\end{figure}

\textbf{Untargeted attacks.}
Un-targeted attacks are `easy' \cite{lord2022attacking} in image classification, especially when the number of classes is large (e.g., in ImageNet that has $1000$ categories). We show that our method can readily achieve a fooling rate over $99\%$ with only $1$--$2$ queries (on average), as depicted in Figure \ref{fig:compare-sota-linf-untargeted} below and Table \ref{tab:pros-cons} in the main text. The initial perturbations from the PM (with all ensemble weights set to $1/N$) can already achieve a fooling rate of over $94\%$, close to that of TREMBA. Other methods require tens or hundreds of queries to achieve near-perfect success rate. 

\section{Experiments on object detection}

To demonstrate the generalizability of BASES beyond classification tasks, we also performed experiments for vanishing attacks on object detectors. The results indicate that our proposed method can be easily adopted for other tasks. 

\subsection{Experiment setup}
\textbf{Surrogate and victim models.} We evaluate BASES using object detectors from MMDetection \cite{mmdetection,code-mmdet}, which provides a diverse set of models form over fifty model families, including \texttt{Faster R-CNN~\cite{ren2015faster}, YOLOv3~\cite{redmon2018yolov3}, RetinaNet~\cite{lin2017focal}, FreeAnchor~\cite{zhang2019freeanchor}, RepPoints~\cite{yang2019reppoints}, CenterNet~\cite{zhou2019objects}, DETR~\cite{carion2020end}, and Deformable DETR~\cite{zhu2021deformable}}. We choose different models \texttt{\{RetinaNet, RepPoints, Deformable DETR\}} as victim blackbox models, as shown in Figure \ref{fig:attack-detectors}. For surrogate models in the PM, we select some popular models \texttt{\{Faster R-CNN, YOLOv3, FreeAnchor, DETR, CenterNet\}} and vary our ensemble size $N \in \{2,3,4,5\}$ by choosing the first $N$ models from the set.

\textbf{Dataset, attacks, query, and perturbation budgets.} All models are trained on COCO 2017 train dataset \cite{lin2014microsoft}. We randomly sample 100 images of stop sign from COCO 2014 validation dataset to perform blackbox vanishing attacks. The attack is considered successful if the victim model fails to detect the stop sign in the adversarial image. The constraints on the query budget $Q \leq 50$ and perturbation budget $\ell_\infty \leq 16$ are the same as the classification setting.

\textbf{Loss functions and ensemble loss.} For individual surrogate models, we use the original loss function used for their training. We defined the ensemble loss as a weighted combination of loss over all the surrogate models. The confidence score of stop sign detected by the victim model is used as a feedback from the victim model. 

\subsection{Attacks on object detection}
\begin{figure}[h]
\centering
\begin{subfigure}[c]{0.325\linewidth}
    \centering
    \includegraphics[width=0.95\textwidth]{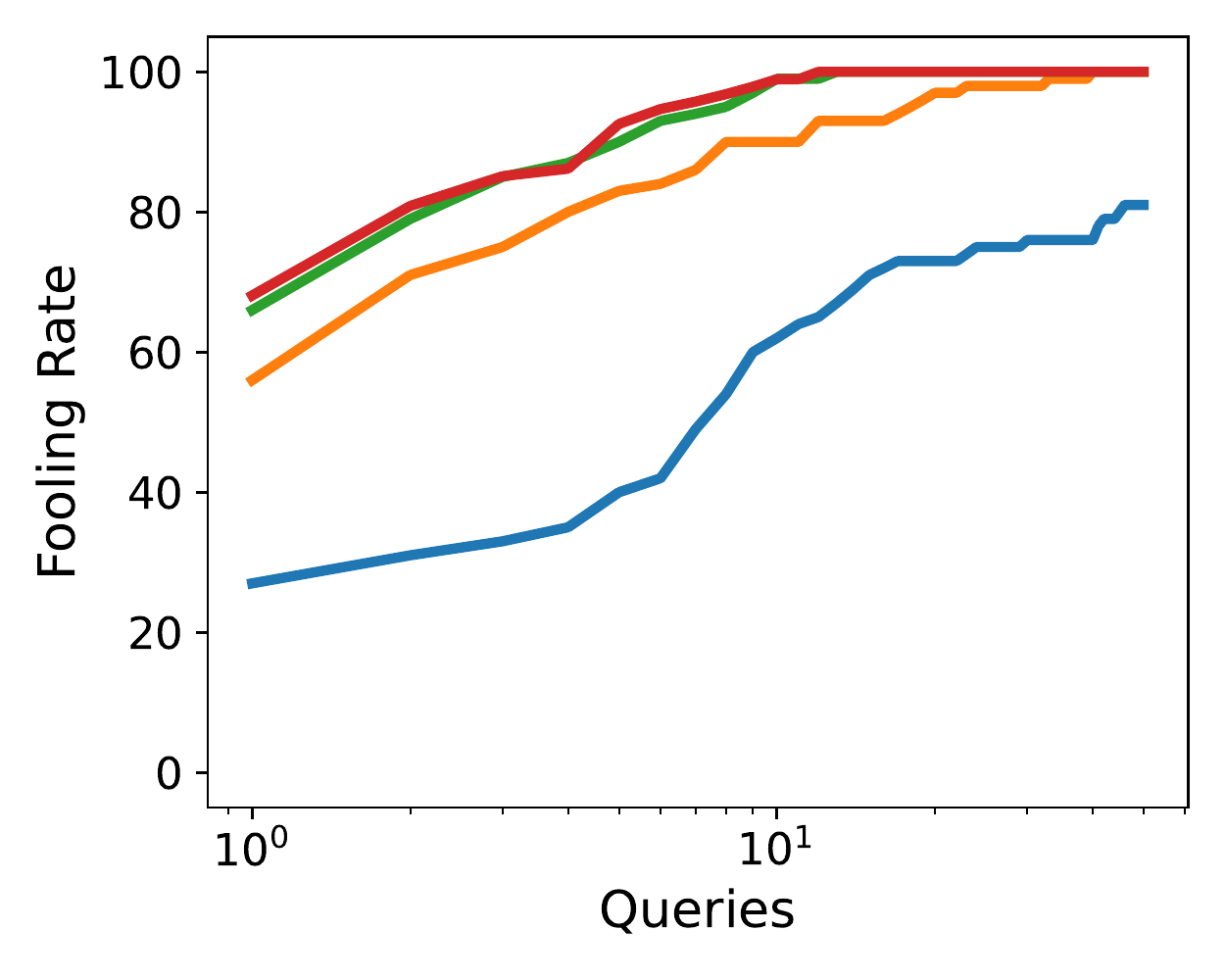}
    \caption{RetinaNet}
\end{subfigure}
\begin{subfigure}[c]{0.325\linewidth}
    \centering
    \includegraphics[width=0.95\textwidth]{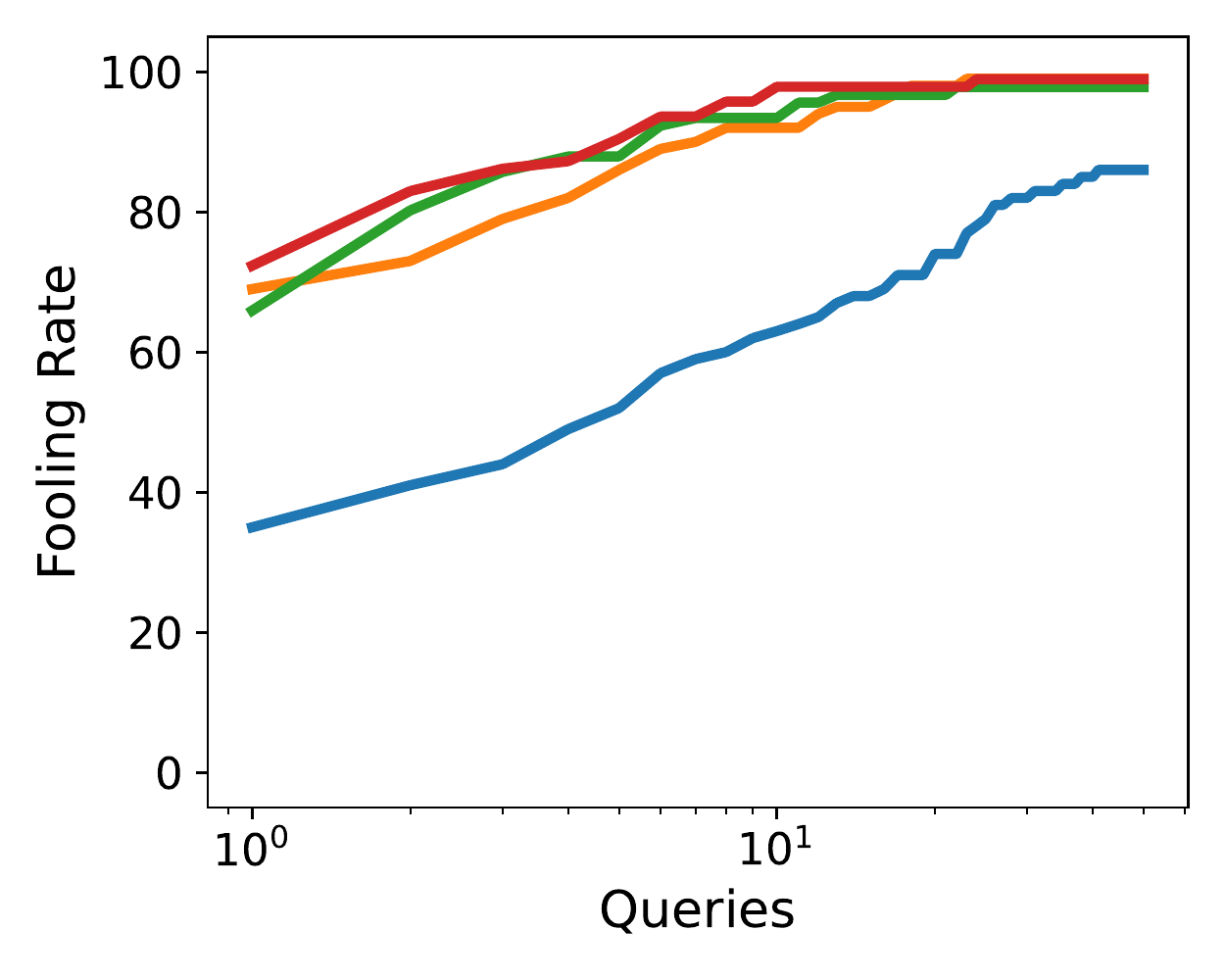}
    \caption{RepPoints}
\end{subfigure}
\begin{subfigure}[c]{0.325\linewidth}
    \centering
    \includegraphics[width=0.95\textwidth]{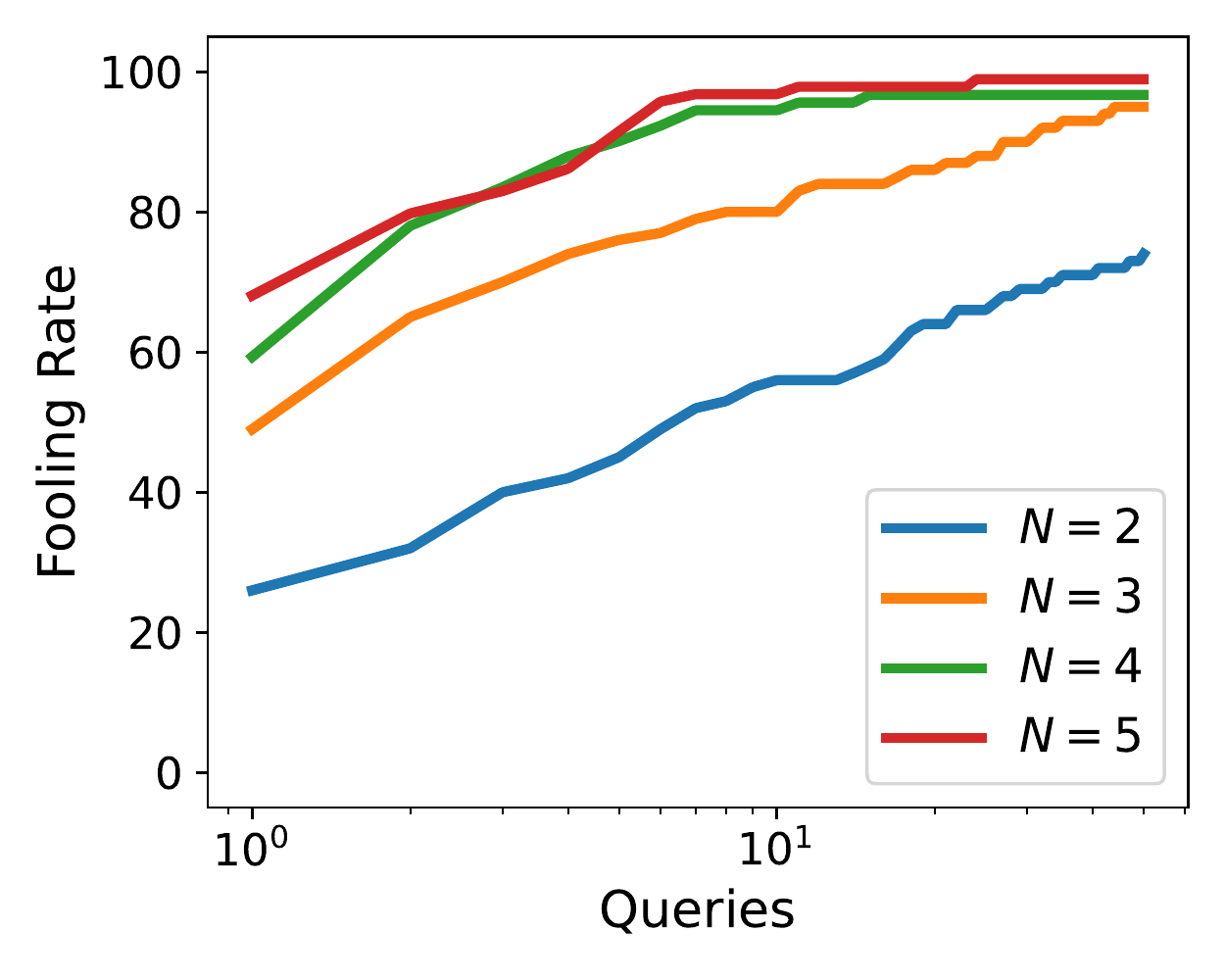}
    \caption{Deformable DETR}
\end{subfigure}
\caption{Fooling rates for vanishing attacks on three victim object detectors using different number ($N \in \{2,3,4,5\}$) of surrogate models in PM.}
\label{fig:attack-detectors}
\end{figure}

The results of attacking object detectors are shown in Figure \ref{fig:attack-detectors} and Table \ref{tab:attack-detectors}. We observe that our attack method is effective and query efficient in attacking object detectors. In particular, for RetinaNet, a simple transfer attack (first iteration) has $27\%$ fooling rate with $N=2$ surrogate models. The fooling rate improve from $27\% \rightarrow 81\%$ with a small number of queries, which is a $300\%$ improvement. Our attack gets stronger as the number of surrogate models increases. When $N=5$, we can get almost perfect ($\geq 99\%$) fooling rate for all victim models with less than 3 queries on average.

\begin{table}[h]
\centering
\small
\caption{Number of queries per image and fooling rate of attacks on three victim models using different number $N$ of surrogate models in PM.}
\label{tab:attack-detectors}
\begin{tabular}{cccc}
\hline
\multirow{2}{*}{$N$} & \multicolumn{3}{c}{Fooling rate and number of queries ($\mathbf{mean}\pm\mathbf{std}$) per image} \\ \cline{2-4} 
 & RetinaNet & RepPoints & Deformable DETR \\ \hline
 \vspace{0.1cm}
2 & 81\% ; 8.5 $\pm$ 11 & 86\% ; 8.0 $\pm$ 9.9 & 74\% ; 8.5 $\pm$ 11 \\
\vspace{0.1cm}
3 & 100\% ; 3.9 $\pm$ 6.5 & 99\% ; 2.8 $\pm$ 4.1 & 95\% ; 5.4 $\pm$ 9.3 \\
\vspace{0.1cm}
4 & 100\% ; 2.2 $\pm$ 2.4 & 98\% ; 2.2 $\pm$ 3.1 & 97\% ; 2.1 $\pm$ 2.2 \\
5 & 100\% ; 2.0 $\pm$ 2.1 & 99\% ; 2.1 $\pm$ 3.0 & 99\% ; 2.1 $\pm$ 2.9 \\ \bottomrule
\end{tabular}
\end{table}

\subsection{Attacks on Google Cloud Vision API}
\begin{figure}[h]
\centering
\begin{subfigure}[c]{0.475\linewidth}
    \centering
    \includegraphics[width=0.95\textwidth]{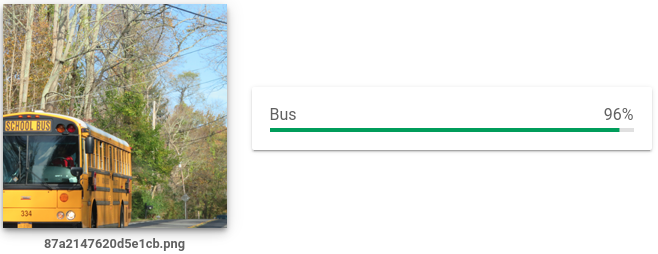}
    \caption{Original Image - Bus}
\end{subfigure}
\begin{subfigure}[c]{0.475\linewidth}
    \centering
    \includegraphics[width=0.95\textwidth]{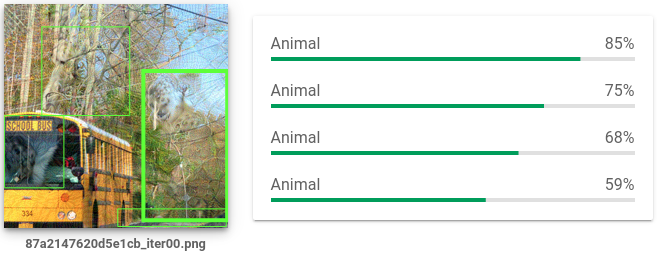}
    \caption{Attacked Image}
\end{subfigure}
\\ \vspace{0.4cm}
\begin{subfigure}[c]{0.475\linewidth}
    \centering
    \includegraphics[width=0.95\textwidth]{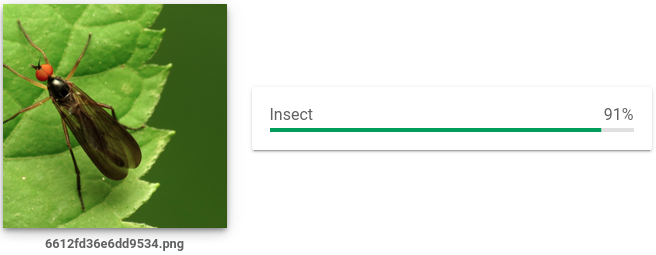}
    \caption{Original Image - Fly}
\end{subfigure}
\begin{subfigure}[c]{0.475\linewidth}
    \centering
    \includegraphics[width=0.95\textwidth]{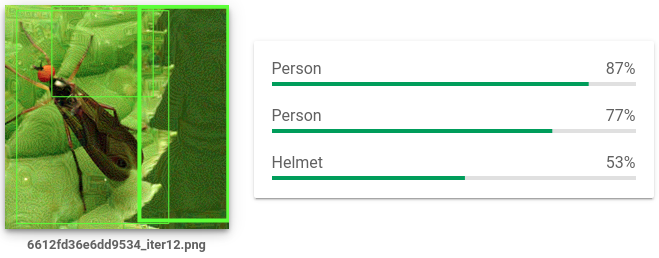}
    \caption{Attacked Image}
\end{subfigure}
\caption{Attacks generated by our PM can fool object detection models. Visualization of some successful attacks on Google Cloud Vision object detection API. (Compare to Figure \ref{fig:gcv-api} in main text.)}
\label{fig:gcv-detector}
\end{figure}

We also observe that the attacks generated by our method can also fool object detection models, as shown in Figure \ref{fig:gcv-detector}.

\clearpage 
\section{Visualization of adversarial examples}

\begin{figure}[h]
\centering
\begin{subfigure}[c]{0.925\linewidth}
    \centering
    \includegraphics[width=0.95\textwidth]{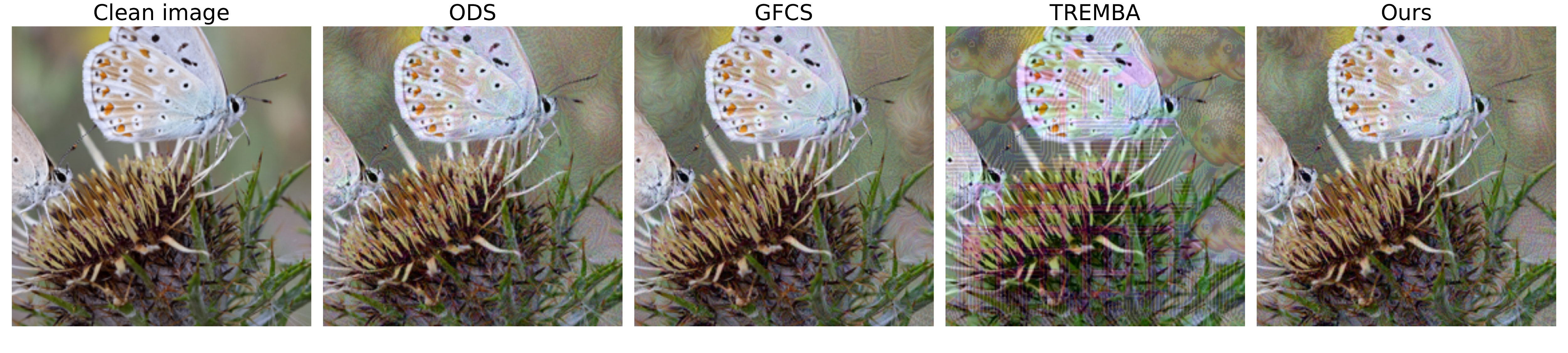}
    \caption{Original: Butterfly; Target: Dog for ODS, GFCS, and Ours; Target: 'Tench' for TREMBA.}
\end{subfigure}
\\ \vspace{0.3cm}
\begin{subfigure}[c]{0.925\linewidth}
    \centering
    \includegraphics[width=0.95\textwidth]{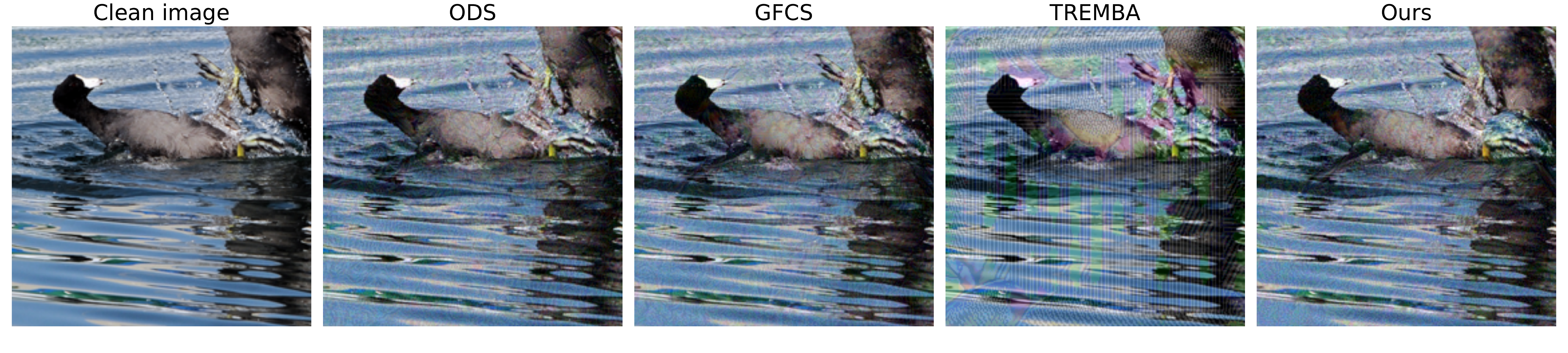}
    \caption{Original: Coot; Target: Jacamar for ODS, GFCS, and Ours; Target: 'Tench' for TREMBA.}
\end{subfigure}
\\ \vspace{0.3cm}
\begin{subfigure}[c]{0.925\linewidth}
    \centering
    \includegraphics[width=0.95\textwidth]{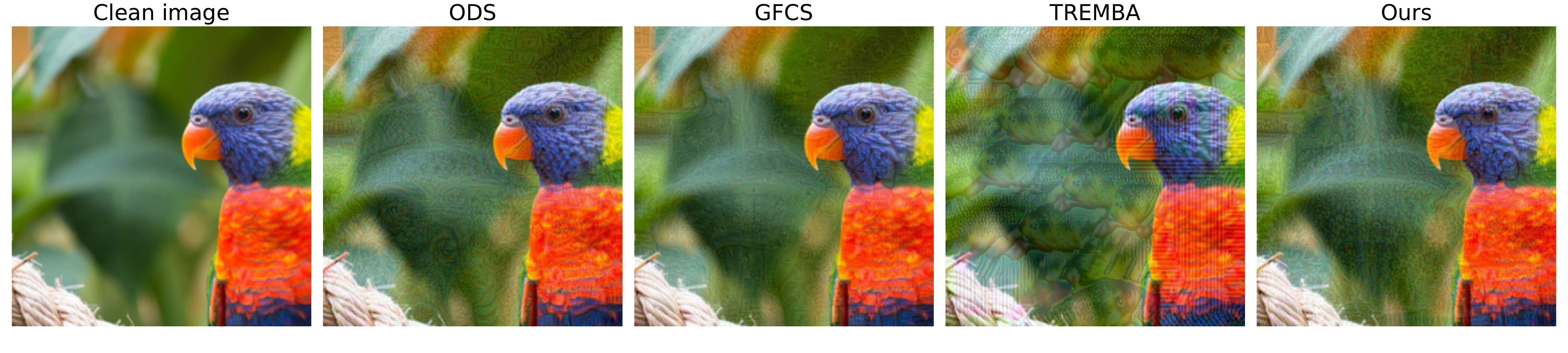}
    \caption{Original: Parrot; Target: Fountain for ODS, GFCS, and Ours; Target: 'Tench' for TREMBA.}
\end{subfigure}
\caption{Visualization of adversarial images generated by different methods for targeted attack. (Corresponds to experiments in Figure \ref{fig:compare-sota-linf-targeted} in main text.)}
\label{fig:visualize-classifiers}
\end{figure}

\textbf{Classifiers}. We present some examples of adversarial images generated by different methods for targeted attack on VGG-19 classifier in Figure \ref{fig:visualize-classifiers}. We observe that even with the same perturbation budget, $\ell_\infty \leq 16$, perturbation from our method is less visible than TREMBA, and is comparable with the ones from ODS and GFCS. TREMBA perturbs all images to `Tench' and has a very structured semantic pattern that becomes visible. ODS, GFCS, and our method perturb `Butterfly' to `Dog', `Coot' to `Jacamar', and `Parrot' to `Fountain'.

\begin{figure}[h]
\centering
\begin{subfigure}[c]{0.925\linewidth}
    \centering
    \includegraphics[width=0.95\textwidth]{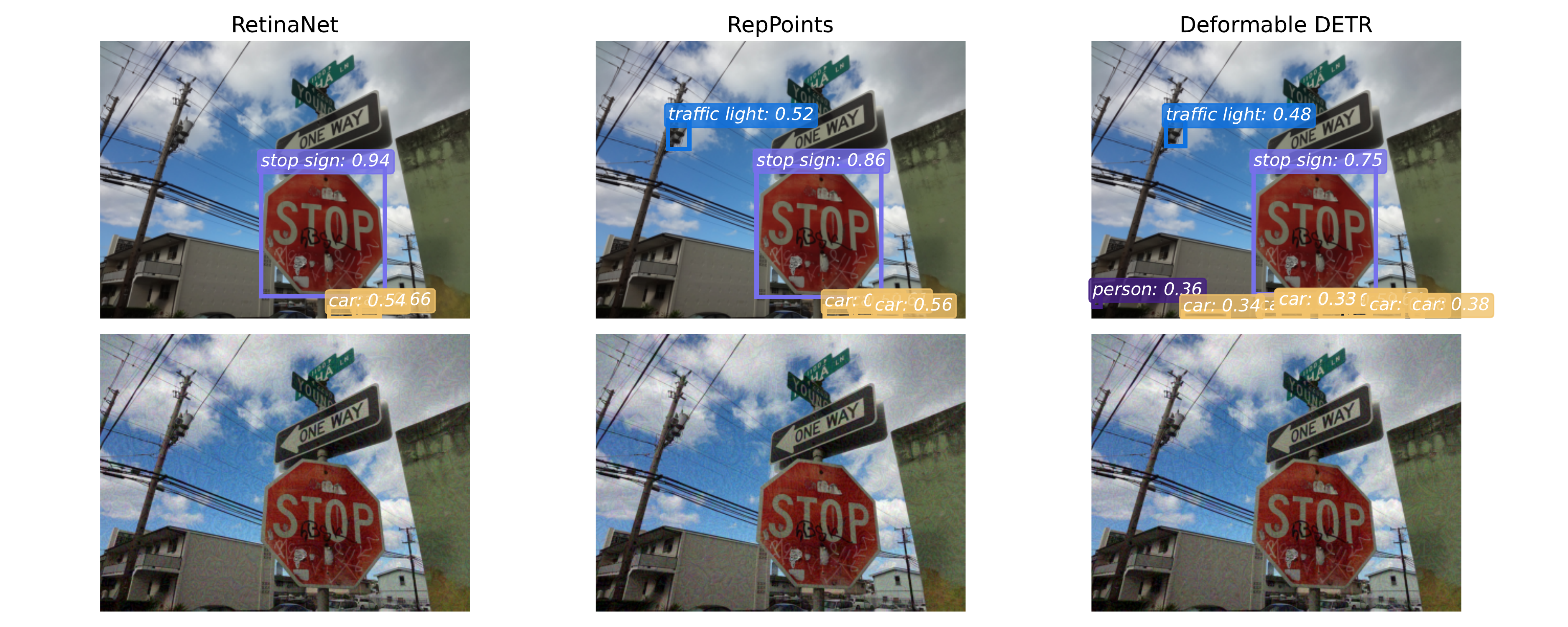}
\end{subfigure}
\\ \vspace{0.2cm}
\begin{subfigure}[c]{0.925\linewidth}
    \centering
    \includegraphics[width=0.95\textwidth]{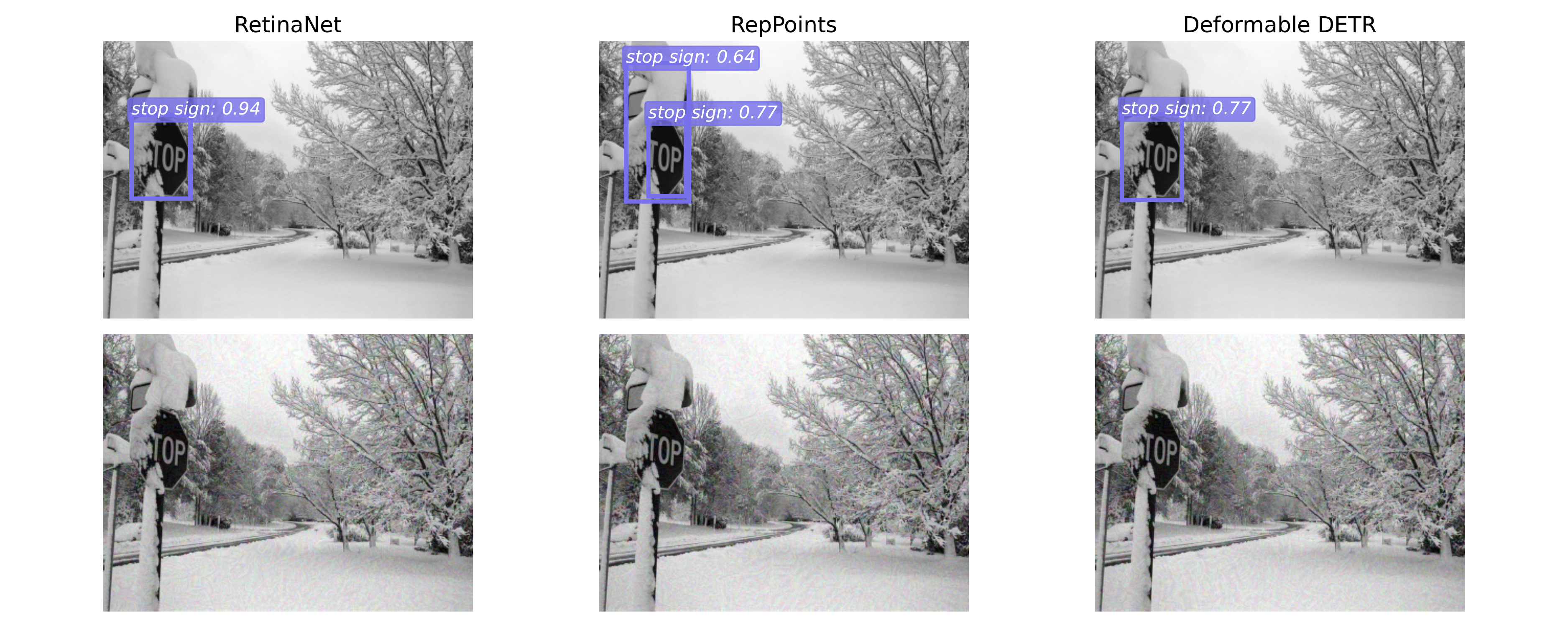}
\end{subfigure}
\\ \vspace{0.2cm}
\begin{subfigure}[c]{0.925\linewidth}
    \centering
    \includegraphics[width=0.95\textwidth]{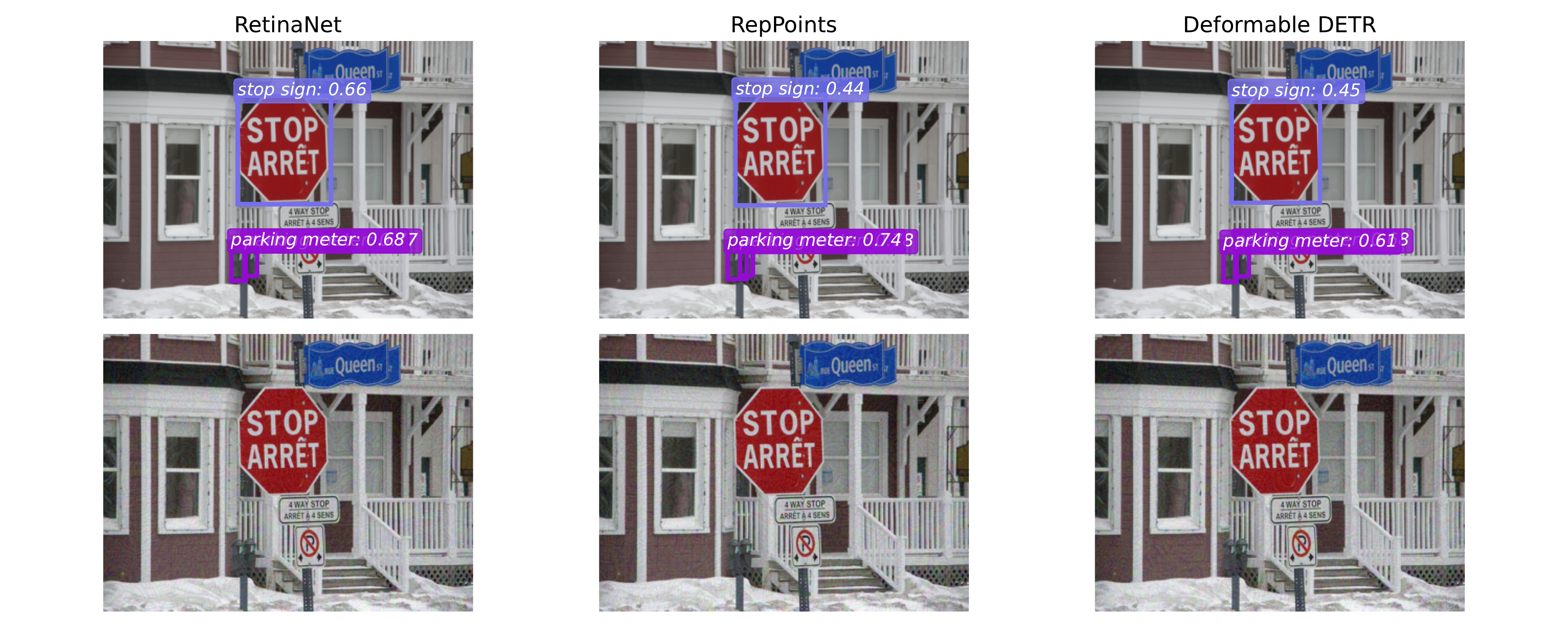}
\end{subfigure}
\caption{Visualization of adversarial images generated by different methods for vanishing attacks on `stop sign'. Top row is detection on clean images and bottom row is detection on adversarial images. (Corresponds to results in Figure \ref{fig:attack-detectors} with $N=5$.)}
\label{fig:visualize-detectors}
\end{figure}

\newpage 
\textbf{Detectors}. We visualize some example images of attacking different object detectors in Figure \ref{fig:visualize-detectors}. Our method effectively vanishes stop sign in the scene.

\clearpage 
\section{Loss landscape vs ensemble weights}
\begin{figure}[h]
\centering
\begin{subfigure}[c]{0.325\linewidth}
    \centering
    \includegraphics[width=0.95\textwidth]{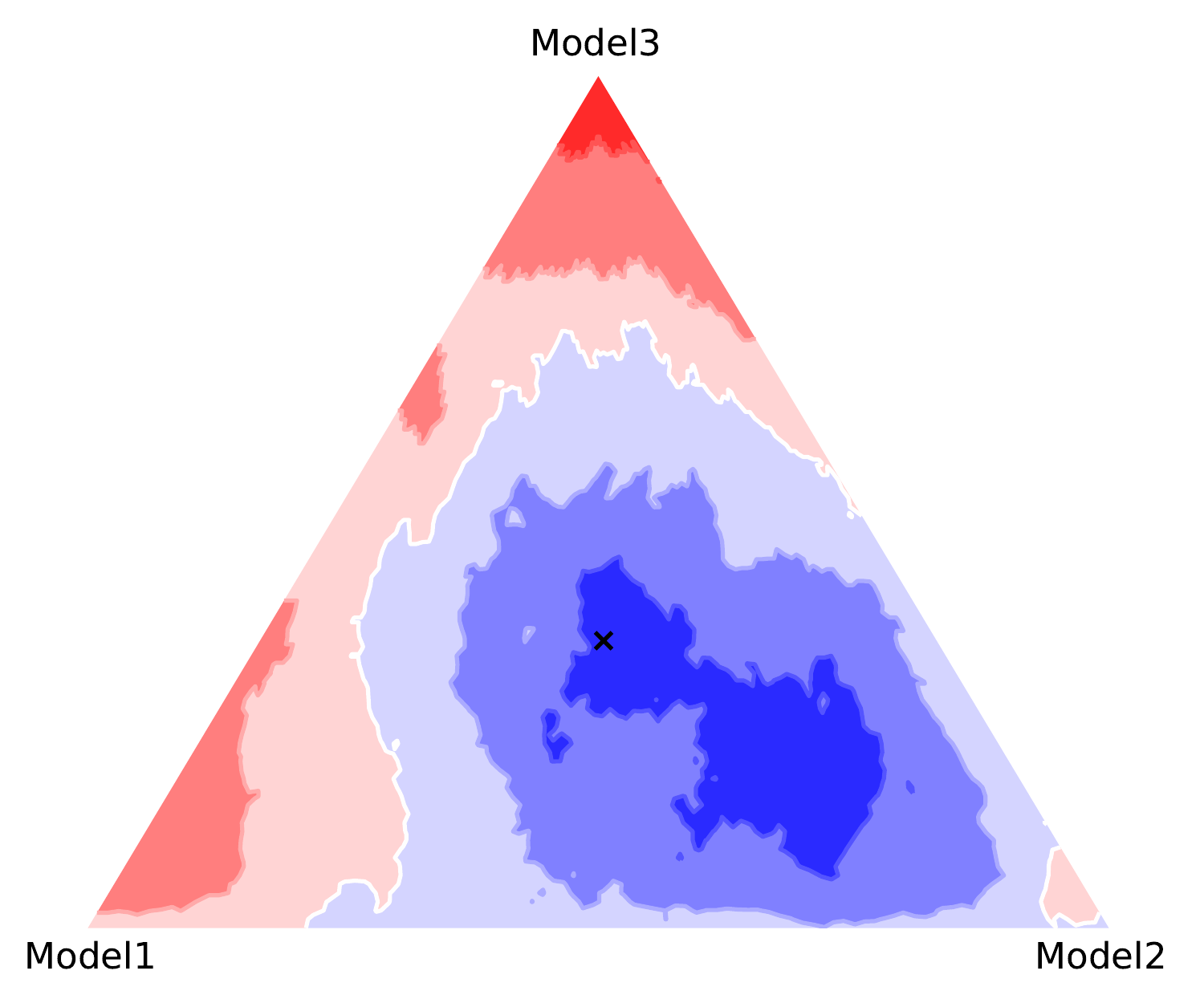}
    \caption{VGG-19, Image A}
\end{subfigure}
\begin{subfigure}[c]{0.325\linewidth}
    \centering
    \includegraphics[width=0.95\textwidth]{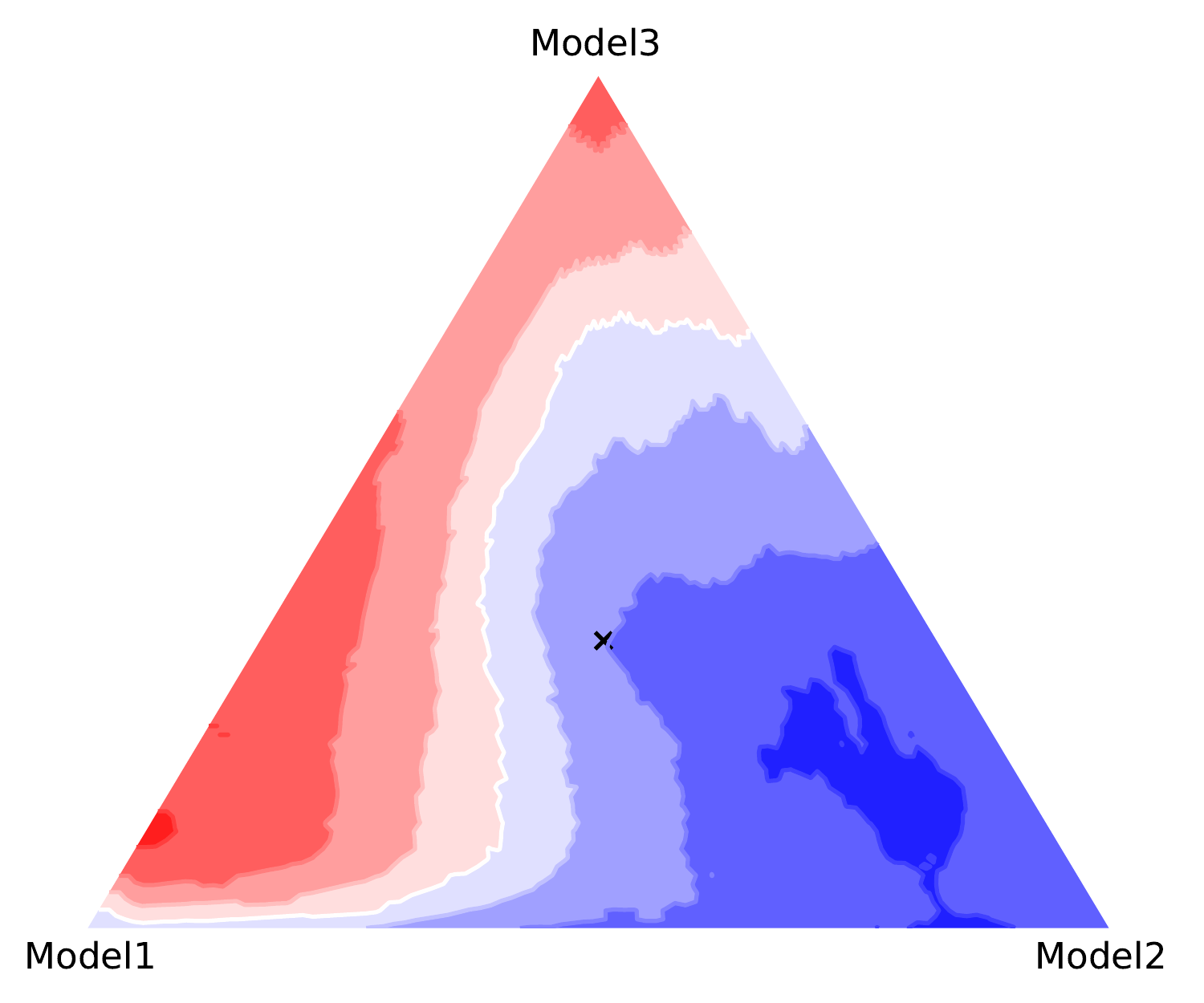}
    \caption{ResNet-34, Image A}
\end{subfigure}
\begin{subfigure}[c]{0.325\linewidth}
    \centering
    \includegraphics[width=0.95\textwidth]{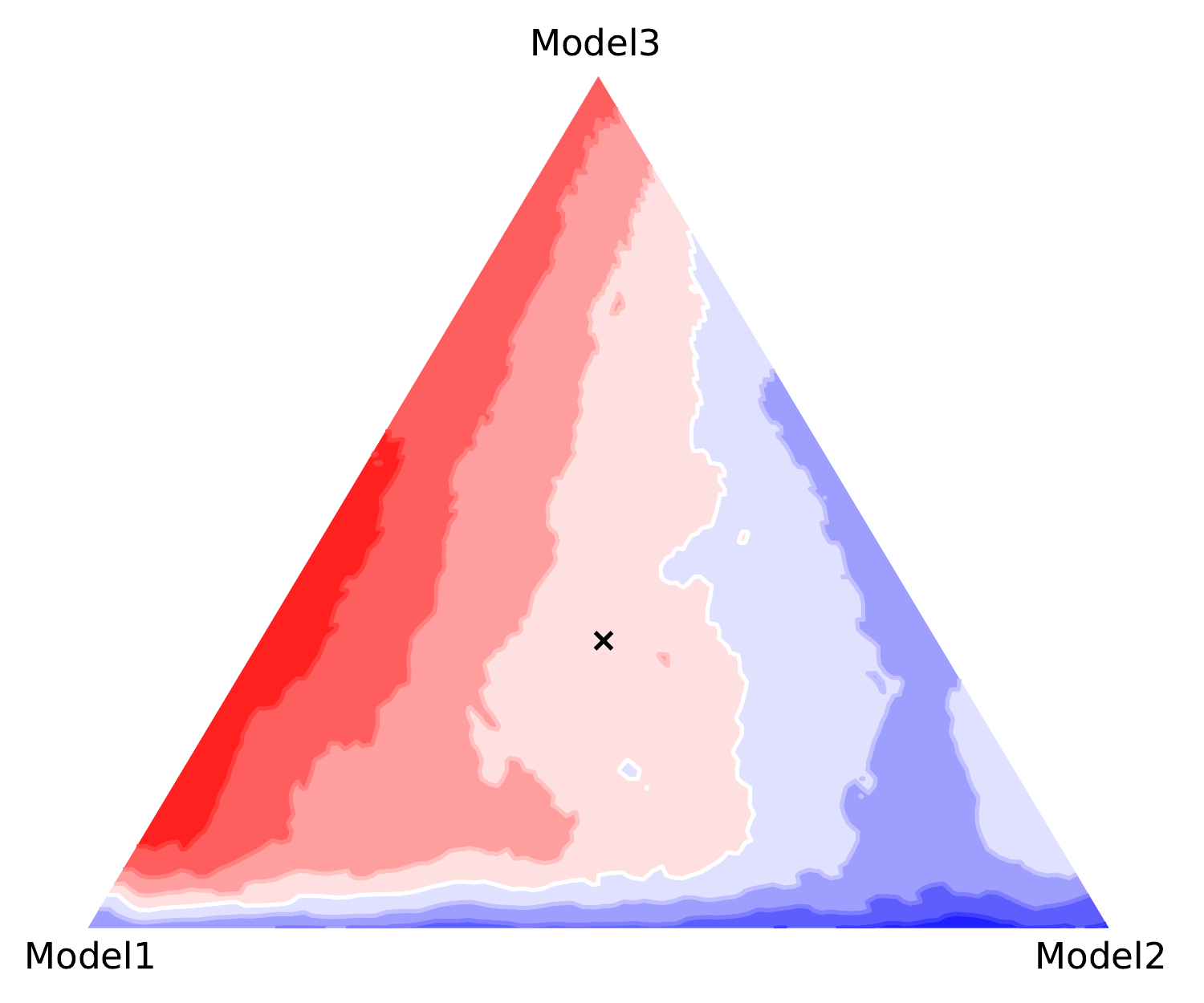}
    \caption{ResNet-34, Image B}
\end{subfigure}
\caption{Illustration of the effect of weights of ensemble models on the attack loss for a victim model. Red color
indicates large loss values (unsuccessful attack), and blue indicates small loss (successful attack).}
\label{fig:illustration-weights}
\end{figure}

\textit{Why does ensemble weights-based query update work?}
We visualize the loss landscape of some victim models with respect to ensemble weights of three surrogate models in the PM. The plots in Figure \ref{fig:illustration-weights} illustrate the loss, where the vertices of each triangle represent the surrogates models in the PM used for attacking a victim model on one image (as shown in sub-caption). The location of each point inside the triangle corresponds to the weight vector $\vw$ (in terms of Barycentric coordinates). For instance, the centroid (marked by $\boldsymbol{\times}$) has the barycentric coordinate $\vw = [\nicefrac{1}{3}, \nicefrac{1}{3}, \nicefrac{1}{3}]$, which implies the losses for all the surrogate models in the ensemble are weighted equally. More weight is given to a model if the weight vector moves closer to the vertex of that model. 
The color of each point inside the triangle represents the victim loss value for the corresponding $\vw$. The attack is more successful when the loss value is low (indicated by blue color) and less successful when the loss value is high (indicated by red color). 
We created this figure using VGG-16, ResNet-18, and SqueezeNet as Model 1,2, and 3, respectively. The main takeaway is that, in many cases, an arbitrary weight vector does not provide successful perturbation for a given victim model; therefore, we need to adjust the weights to generate successful attacks. 

\section{Classification performance on clean images}
To ensure that all the models provide reasonably correct  classification results on clean images, we calculate the classification accuracy of all ImageNet models on the 1000 test images. Our calculation shows that they have a Top-1 accuracy of ($\mathbf{mean}\pm\mathbf{std}$): $89.1\% \pm 6.5\%$. Among all the models tested, \texttt{Convnext-Smal} achieves the highest accuracy at $96.8\%$, and \texttt{SqueezeNet-1.1} gets the lowest at $68.8\%$. 
\begin{figure}[h]
\centering
\begin{subfigure}[c]{\linewidth}
    \centering
    \includegraphics[width=\textwidth]{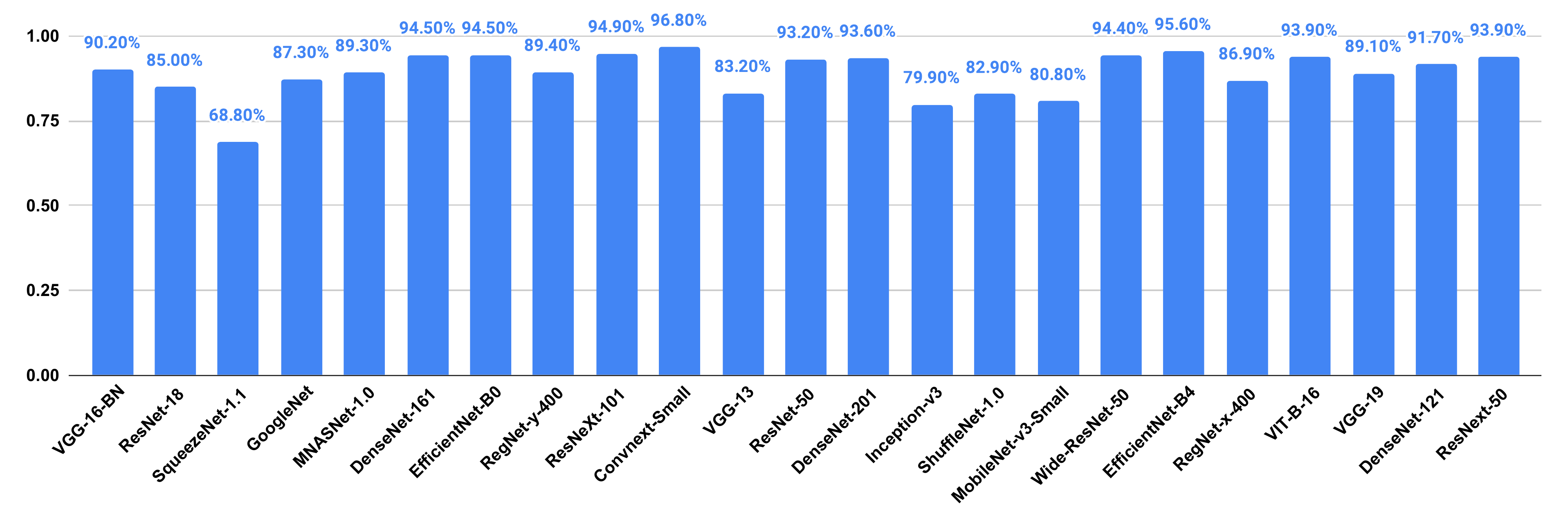}
\end{subfigure}
\caption{Top 1 classification accuracies of different ImageNet models used in our experiments.}
\label{fig:clean-acc}
\end{figure}

\end{document}